\documentclass[journal]{IEEEtran}
\usepackage{amsmath,amsfonts,amssymb}
\usepackage{algorithmic}
\usepackage{array}
\usepackage[caption=false,font=normalsize,labelfont=sf,textfont=sf]{subfig}
\usepackage{textcomp}
\usepackage{stfloats}
\usepackage{url}
\usepackage{verbatim}
\def\BibTeX{{\rm B\kern-.05em{\sc i\kern-.025em b}\kern-.08em
    T\kern-.1667em\lower.7ex\hbox{E}\kern-.125emX}}
\usepackage{balance}
\usepackage{booktabs}
\usepackage{graphicx}
\usepackage{multirow}
\usepackage{ntheorem}
\usepackage{bbding}
\usepackage[colorlinks,anchorcolor=blue]{hyperref}
\usepackage{algorithm}
\usepackage{algorithmic}
\usepackage{bm}

\theorembodyfont{\rmfamily}

\newtheorem{definition}{Definition}
\newtheorem{theorem}{Theorem}
\newtheorem{prop}{Proposition}

\newcolumntype{L}{>{\centering\arraybackslash}m{.5cm}}

\newcolumntype{S}{>{\centering\arraybackslash}m{.65cm}}

\DeclareMathOperator*{\argmax}{\arg\!\max}

\begin{document}

\title{Partial Domain Adaptation via Importance Sampling-based Shift Correction}

\author{Cheng-Jun Guo,~Chuan-Xian Ren,~You-Wei Luo,~Xiao-Lin Xu, and~Hong~Yan~\IEEEmembership{Fellow,~IEEE} 
\IEEEcompsocitemizethanks{\IEEEcompsocthanksitem C.J. Guo, C.X. Ren, and Y.W. Luo are with the School of Mathematics, Sun Yat-Sen University, Guangzhou, 510275, China. X.L. Xu is with the School of Statistics and Mathematics, Guangdong University of Finance and Economics, Guangzhou 510320, China. H. Yan is with Department of Electrical Engineering, City University of Hong Kong, 83 Tat Chee Avenue, Kowloon, Hong Kong. C.X. Ren is the corresponding author (rchuanx@mail.sysu.edu.cn).
\IEEEcompsocthanksitem This work is supported in part by National Natural Science Foundation of China (Grants No. 62376291), in part by Guangdong Basic and Applied Basic Research Foundation (2023B1515020004), in part by Science and Technology Program of Guangzhou (Grants No. 2024A04J6413), in part by the Fundamental Research Funds for the Central Universities, Sun Yat-sen University (24xkjc013), in part by the Hong Kong Innovation and Technology Commission (InnoHK Project CIMDA), and in part by the Institute of Digital Medicine, City University of Hong Kong (Project 9229503).
\IEEEcompsocthanksitem This paper has supplementary downloadable material available at http://ieeexplore.ieee.org., provided by the author.}}

\markboth{Journal of \LaTeX\ Class Files,~Vol.~18, No.~9, September~2020}%
{How to Use the IEEEtran \LaTeX \ Templates}

\maketitle

\begin{abstract}
Partial domain adaptation (PDA) is a challenging task in real-world machine learning scenarios. It aims to transfer knowledge from a labeled source domain to a related unlabeled target domain, where the support set of the source label distribution subsumes the target one. Previous PDA works managed to correct the label distribution shift by weighting samples in the source domain. However, the simple reweighing technique cannot explore the latent structure and sufficiently use the labeled data, and then models are prone to over-fitting on the source domain. In this work, we propose a novel importance sampling-based shift correction (IS$^2$C) method, where new labeled data are sampled from a built sampling domain, whose label distribution is supposed to be the same as the target domain, to characterize the latent structure and enhance the generalization ability of the model. We provide theoretical guarantees for IS$^2$C by proving that the generalization error can be sufficiently dominated by IS$^2$C. In particular, by implementing sampling with the mixture distribution, the extent of shift between source and sampling domains can be connected to generalization error, which provides an interpretable way to build IS$^2$C. To improve knowledge transfer, an optimal transport-based independence criterion is proposed for conditional distribution alignment, where the computation of the criterion can be adjusted to reduce the complexity from $\mathcal{O}(n^3)$ to $\mathcal{O}(n^2)$ in realistic PDA scenarios. Extensive experiments on PDA benchmarks validate the theoretical results and demonstrate the effectiveness of our IS$^2$C over existing methods.
\end{abstract}

\begin{IEEEkeywords}
    Partial domain adaptation, importance sampling, generalization error analysis, label shift, conditional shift.
\end{IEEEkeywords}

\section{Introduction}

\IEEEPARstart{S}{upervised} learning has made remarkable advances across a wide variety of machine learning and pattern recognition tasks. Sufficient labeled data plays an important role in supervised learning for training a well-performing model. However, not all tasks have sufficient labeled data for model training, and collecting enough labeled data is commonly at a great cost of time and resources. To relieve the burden of expensive data labeling, transfer learning has been introduced to transfer useful knowledge from the existing labeled training (source) data to the related unlabeled test (target) data. Unfortunately, the source and target data are generally from different domains, and there exists a large distribution discrepancy between different domains due to a variety of factors:  different corrections for atmospheric scattering, daylight conditions at the hour of acquisition, or even slight changes in the chemical composition of the materials \cite{courty2017optimal}. Therefore, directly applying a well-trained source domain model on the target domain inevitably leads to severe degradation of classification performance.

\begin{figure}[t] \centering
    \includegraphics[scale=0.329]{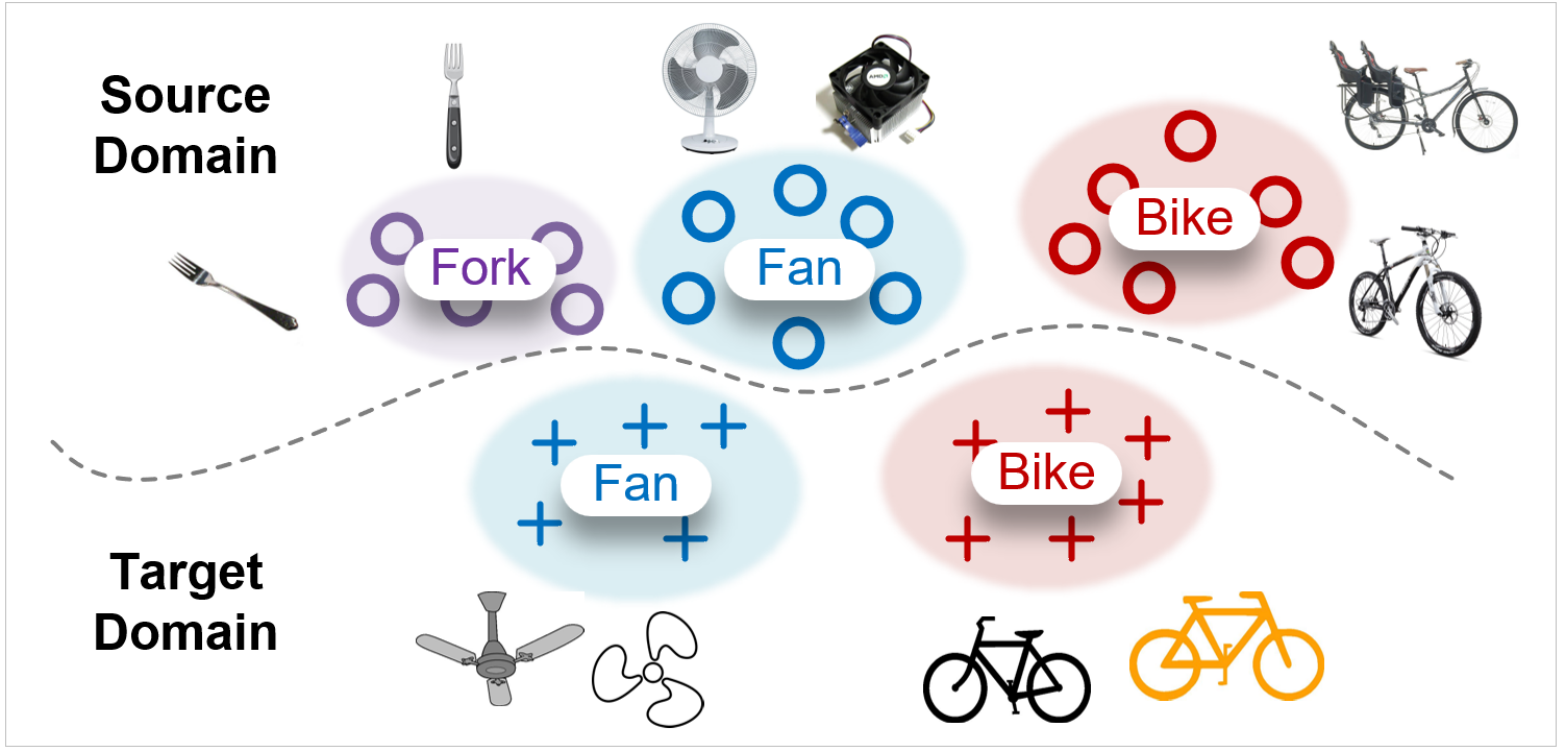}\hspace{0in}
    \caption{Partial domain adaptation, in which the support of source label distribution subsumes the target one. This scenario is difficult since there may exist outlier classes in the source domain, i.e., classes not appearing in the target domain, e.g. the ‘fork’ class. The outliers belonging to these classes would affect the model training procedure and harm the model performance on the target domain. In this paper, we propose a novel method, i.e., importance sampling-based shift correction, to address the problems.}
    \label{fig:PDA}
\end{figure}

Unsupervised domain adaptation (UDA) has been proposed to mitigate the degradation of model performance caused by domain gap~\cite{pan2010domain, liang2021SHOT}. In UDA, the source and target domains are assumed to have similar but not identical distributions $P_s$ and $P_t$, where the supports of the label distributions $P_s(Y)$ and $P_t(Y)$ are the same. To make the problem more practical, most UDA methods generally assume covariate shift, where conditional distributions of the labels with respect to the samples are equal (i.e., $P_s(Y|X)=P_t(Y|X)$), while the marginal distributions in source and target domains are different (i.e. $P_s(X)\neq P_t(X)$). Following this line, theoretical results are derived by Ben-David et al. \cite{ben2010theory,david2010impossibility}, and empirical methods are explored by mitigating marginal distribution shift via maximum mean discrepancy (MMD)~\cite{borgwardt2006integrating, long2015learning, zhu2020DSAN}, covariance matching~\cite{sun2016return}, subspace alignment~\cite{gong2012geodesic, ren2019heterogeneous, luo2020unsupervised}, adversarial training~\cite{ganin2016domain,tzeng2017adversarial,tang2020discriminative,ren2019domain} and optimal transport (OT)~\cite{courty2017joint,damodaran2018deepjdot,zhang2019optimal}. However, these UDA methods implicitly assume that the distributions across domains are similar enough so that the source classes will not be mixed with irrelevant target classes, which is hard to satisfy in real-world applications. Though several methods try to achieve class-wise alignment by incorporating label information into adaptation, the misalignment cannot be avoided. 

To address the limitation of UDA setting, partial domain adaptation (PDA) is proposed by relaxing the assumption on label variables. Specifically, PDA assumes the label distribution changes across domains, where the extreme scenario of such a label shift setting~\cite{ren2018generalized,kirchmeyer2021mapping,tachet2020domain,zhang2013domain,luo2022generalized} is considered, i.e., the proportions of some classes in target domains are $0$ as shown in Figure \ref{fig:PDA}. Most previous PDA methods are built on importance weighting (IW) ~\cite{cao2019learning,li2020deep} strategy, which tries to learn an importance weight $\bm{w}$ and gain a reweighed source $P_s^{\bm{w}}$ for label distribution correction, then align the reweighed source with target distribution from the view of covariate shift \cite{yan2017mind,cao2018partial,kim2020associative}. 

Intuitively, the importance weighting strategy can be understood as assigning weights to source domain samples to mitigate the impact of outliers (those samples belonging to outlier classes) on the model during training. However, the weighting-based model relies heavily on the high quality of the learned weights, since all the outliers with small but non-zero learned weights will still be utilized in the training process and their harmful effect on the model cannot be removed. Reviewing the initial motivation for studying transfer learning, the core issue is the lack of labeled data. Therefore, in PDA, if there are sufficient labeled data available to form a sampling domain that has a distribution approximating the target distribution, then sampling is a more practical way than weighting. Ideally, the sampling data would reduce the impacts of outliers as they are removed, and also contribute to better generalization of the learned model.

Based on the motivation above, we propose a novel importance sampling-based shift correction (IS$^2$C) method, which directly samples new labeled data from a built sampling domain to train a model and explicitly aligns the class-conditional distributions on shared classes between the source and target domains, i.e., aligns $P_s(g(X)|Y=j)$ with $P_t(g(X)|Y=j)$ where $p_t(Y=j)\neq0$, and $g$ is a feature transformation to be learned. To provide a better understanding on the working mechanism of IS$^2$C, we further explore the theoretical guarantee for IS$^2$C. Our results show that the target classification risk can be sufficiently dominated by IS$^2$C. Further, we propose a theory-driven learning model for dealing with PDA, which consists of 1) an importance sampling (IS) module to correct the shifts on label space; 2) the OT-based independence criterion, i.e., entropy regularized optimal transport independence criterion (ETIC), for learning representations with aligned class-conditional distributions, where a fast computation formulation is derived to speed up the model efficiency in practical applications.

Our main contributions can be summarized as follows.
\begin{enumerate}
    \item A novel importance sampling-based correction method is proposed for PDA, which characterizes the latent structure and improves model generalization ability by sampling new data that are unseen in the original space and learning conditional independent representations for cross-domain data.
    \item Theoretical guarantees for IS$^2$C are explored: (i) an informative generalization risk analysis for the sampling domain is presented, which proves that the risk can be sufficiently bounded by IS$^2$C; (ii) the learning model on the sampling domain admits smaller task risk than the original domain, which ensures IS$^2$C can further reduce the generalization risk.
    \item A practical learning model is proposed under the IS$^2$C. Specifically, to improve knowledge
transfer, an ETIC-based approach is proposed to mitigate class-conditional discrepancy, where the computation of ETIC can be adjusted to reduce the complexity from $\mathcal{O}(n^3)$ to $\mathcal{O}(n^2)$ in realistic PDA scenarios.
\end{enumerate}

The rest of this paper is organized as follows. In Section \ref{sec2}, we provide a brief review of UDA and PDA, followed by a detailed explanation of our IS$^2$C and theoretical analysis in Section \ref{sec3}. In Section \ref{sec4}, we present the algorithm for practical PDA applications, including the ETIC-based alignment loss, the fast computation of ETIC, and the final optimization objective. Extensive experiments and analysis under PDA scenarios are presented in Section \ref{sec5}. Finally, Section \ref{sec6} concludes this paper.

\section{Related Work}\label{sec2}
The classical learning paradigm assumes that $ P_s(X,Y) $ is the same as $ P_t(X,Y) $, so the learned model can be directly applied to the unlabeled target data. However, in the practical UDA setting, the identical distribution assumption on cross-domain distributions is relaxed, where the main goal is to tackle the discrepancy between $ P_s(X,Y) $ and $ P_t(X,Y) $ \cite{pan2010domain}.

To address the distribution shift issue, pioneering works are developed by minimizing the discrepancy between domains, e.g., moment matching \cite{long2015learning,sun2016return,ge2023unsupervised}, manifold-based method~\cite{luo2020unsupervised}, adversarial-based method \cite{ganin2016domain,long2018conditional,xu2020adversarial}, and OT-based method \cite{courty2017optimal,li2020enhanced,luo2021conditional,ren2022buresnet}. Specifically, moment matching methods aim to align the first-order or second-order statistics for transfer, e.g., deep adaptation network~\cite{long2015learning}, deep conditional adaptation network~\cite{ge2023unsupervised}, correlation alignment~\cite{sun2016return}. Manifold-based methods focus on the subspace alignment via norm criterion or Riemannian metric, e.g., adaptive feature norm (AFN) \cite{xu2019larger} and discriminative manifold propagation (DMP) \cite{luo2020unsupervised}. Adversarial-based methods learn domain-invariant representations by training a domain-indistinguishable feature extractor and domain discriminator, e.g., domain-adversarial neural network (DANN)~\cite{ganin2016domain}, conditional domain adversarial network~\cite{long2018conditional}. To further improve the robustness of the model, Xu et al.~\cite{xu2020adversarial} leverage the mixup technique to mix samples from the source and target domains for adversarial domain adaptation. By introducing the mixup technique to form a vicinal space, Na et al.~\cite{na2022contrastive} propose a contrastive vicinal space-based algorithm, which fully utilizes vicinal instances from the perspective of self-training to mitigate domain discrepancy. OT-based methods propose to decrease the domain discrepancy by minimizing total transport cost, where Euclidean distance~\cite{courty2017optimal}, enhanced transport distance~\cite{li2020enhanced}, conditional kernel Bures metric~\cite{luo2021conditional,ren2022buresnet}. However, in real-world scenarios, there is always a discrepancy between label distributions, i.e., label/target shift. Since these methods typically assume identical label spaces for source and target domains, they often suffer from negative transfer, resulting in severe performance degradation.

PDA can be seen as an extreme scenario of label shift, where the proportions of some classes in target domains are $0$, i.e., $ {\rm supp}(P_t(Y))\subsetneq {\rm supp}(P_s(Y)) $. A natural idea to address PDA is to correct the label shift and turn PDA to UDA. Inspired by such an idea, many PDA methods have been developed under importance weighting strategy, e.g., reweighed moment matching \cite{yan2017mind}, reweighed adversarial method \cite{cao2018partial}, and reweighed OT method \cite{gu2021adversarial,nguyen2022improving,luo2023mot}.  Sahoo et al. \cite{sahoo2023select} propose a strategy to learn discriminative invariant feature representations for partial domain adaptation through select, label, and mix (SLM) modules. Similarly, Wu et al.~\cite{wu2022ran} introduce a selection module based on a deep reinforcement learning model and propose the reinforced adaptation
network (RAN). Cao et al.~\cite{cao2022san} uncover the importance of estimating the transferable probability of each instance and propose the selective adversarial network (SAN++). Li et al.~\cite{li2022idsp} attempt to ignore non-identity between support sets of label distributions by giving up possibly riskier domain alignment, and propose the intra-domain structure preserving (IDSP) method for PDA. Weighted domain adaptation network~\cite{yan2017mind} takes label distributions of the source and target domains into account, and it then introduces the reweighed source domain to compute the weighted maximum mean discrepancy. To ensure convergence in training, Ma et al. \cite{ma2024small} propose soft-weighted maximum mean discrepancy and use it instead of the weighted maximum mean discrepancy to minimize the cross-domain distribution discrepancy. Partial adversarial domain adaptation (PADA) \cite{cao2018partial} averages the label predictions to calculate the weight $\bm{w}$ to obtain the reweighed source domain, and then trains a model on the reweighed source as DANN. Mini-batch partial optimal transport (m-POT) \cite{nguyen2022improving} uses the relaxation-based OT distance without strict marginal constraints, which is a better metric for domain discrepancy in PDA. Luo et al. \cite{luo2023mot} incorporate the mask mechanism into unbalanced OT distance and propose masked optimal transport to measure and reduce the conditional discrepancy between the reweighed source and target distributions. Adversarial reweighing (AR) \cite{gu2021adversarial} adversarially learns the weights of source domain data by minimizing the OT distance between the reweighed source distribution and target distribution, and the transferable deep recognition network is learned on the reweighed source domain data. Lin et al.~\cite{lin2022CI} exploit the cycle inconsistency (CI) to learn accurate weight assignment for weight suppression of the outliers, and implement adversarial-based domain alignment. 

Most of these PDA methods focus on learning an importance weight $\bm{w}$ to reweigh the source domain, such that the reweighed source has the same label distribution as the target. In particular, they learn a weight for source samples, and then use the weight and source data to train a model, ignoring the latent structure hidden in the source domain.

Different from the previous methods, our novel IS$^2$C method does not weigh source data. We try to directly sample new labeled data from a sampling distribution $P_c$, and use the sampling data to train a more generalized model. Theoretically, we also derive a generalization error bound to ensure the rationality of IS$^2$C.

\section{IS$^2$C: Theoretical analysis}\label{sec3}

We focus on the $K$-class classification problem. Let $ \mathcal{X}\subset \mathbb{R}^d $ and $ \mathcal{Y}=[K] $ be the input/sample space and output/label space, respectively. Denote $ X $ as the random variable which takes values in $ \mathcal{X} $, and $ Y $, $ \hat{Y} $ are random variables which take values in $ \mathcal{Y} $, where $Y$ represents real label and $\hat{Y}$ represents the predicted label. $ P_s $ and $ P_t $ denote the distributions of source and target domains, e.g., $ P_s(X,Y) $ is the joint distribution of input and real label of the source domain. In this work, we consider the PDA setting, where ${\rm supp}(P_s(Y))=\mathcal{Y}$ and $ {\rm supp}(P_t(Y))\subsetneq \mathcal{Y}$. A model $h\circ g$ consists of a feature transformation $ g(\cdot):\mathcal{X}\to \mathcal{F} $, where $\mathcal{F}$ is a feature space induced by $g(\cdot)$, and a classifier $ h(\cdot) $. Classifier $h(\cdot)$ finally outputs a $K$-dimensional probabilistic vector, which represents the predicted distribution $P(\hat{Y}|X)$. The error of a model defined as $ \varepsilon(h\circ g)=\sum_{i\neq j} p(\hat{Y}=i, Y=j) $.

\subsection{Importance Sampling-based Framework}

In this section, we elaborate on the IS$^2$C. First, we give a brief introduction to importance sampling (IS).

IS usually refers to a collection of Monte Carlo methods where a mathematical expectation with respect to a probability density $p(\bm{x})$ is approximated by a weighted average of random draws from another known distribution $q(\bm{x})$ \cite{tokdar2010importance}. Suppose we want to estimate an expectation of a function $f$ over the random variable $X$, i.e.,
\begin{align*}
    \mu_f =\int f(\bm{x})p(\bm{x})d\bm{x}.
\end{align*}
Then for any probability density $q(\bm{x})$ that satisfies $ {\rm supp}(f\cdot p) \subseteq {\rm supp}(q) $, the expectation $\mu_f$ can be rewritten as
\begin{align*}
    \mu_f =\int \frac{p(\bm{x})}{q(\bm{x})}f(\bm{x})q(\bm{x})\,d\bm{x},
\end{align*}
where $\frac{p(\bm{x})}{q(\bm{x})}$ is weight function, $q(\bm{x})$ is distribution for sampling. Therefore, $\mu_f$ can be estimated by independently drawing sample $\{\bm{x}_i\}_{i=1}^m$ from $q(\bm{x})$ as
\begin{align*}
    \mu_f =\frac{1}{m}\sum_{i=1}^m \frac{p(\bm{x}_i)}{q(\bm{x}_i)}f(\bm{x}_i).
\end{align*}

Usually, both $p$ and $q$ are accessible distributions. The reason for introducing distribution $q$ is that it can be challenging to estimate the expectation by sampling from distribution $p$, i.e., the results obtained from sampling distribution $p$ have higher variance, while the results obtained from sampling distribution $q$ have lower variance.

Following the idea of IS, we consider introducing a sampling domain with distribution $P_c$ since the model trained on the original source domain $P_s$ may perform poorly on the target domain $P_t$ due to the negative effect of outliers. A natural idea is that the sampling domain should have the same distribution as the target domain. However, this is not feasible since sufficient labeled data cannot be obtained from the target domain distribution. Therefore, we consider a sampling domain with the same label distribution as the target domain and the same class-conditional distribution as the source domain, while reducing the class-conditional discrepancy between the source and target domains to approximate $P_t$ by $P_c$. Moreover, to further avoid the effect of outliers and the overlap between different classes, a mixture instead of the original source conditional distribution is used so that the cluster structure of each class is more compact while preserving feature diversity. The IS$^2$C is defined as follows.
\begin{definition}[Importance Sampling-based Shift Correction]\label{Def_I2SC}
    Let $g(\cdot)$ be a feature transformation and $ h(\cdot) $ be a classifier.\\
    \qquad (1) For $\forall j\in \mathcal{Y}$, consider $X_1$, $X_2\sim P_s(X|Y=j)$ and $\theta \in [0,1]$. Then a sampling domain with distribution $P_c$ is defined as $p_c(Y=j)=p_t(Y=j)$ and $X=\theta X_1+(1-\theta)X_2$ for $X\sim P_c(X|Y=j)$;  \\
    \qquad (2) Let $g^{\star}$, $h^{\star}=\arg\min_{g,h} ~ \varepsilon_c(h\circ g)$. A model $ h\circ g $ is called a IS$^2$C learning model if $g=g^{\star}$, $h=h^{\star}$ and $P_s(g(X)|Y=j)=P_t(g(X)|Y=j)$ for any $j\in {\rm supp} (P_t(Y))$.
\end{definition}

\begin{figure}[!t] \centering
    \includegraphics[scale=0.325]{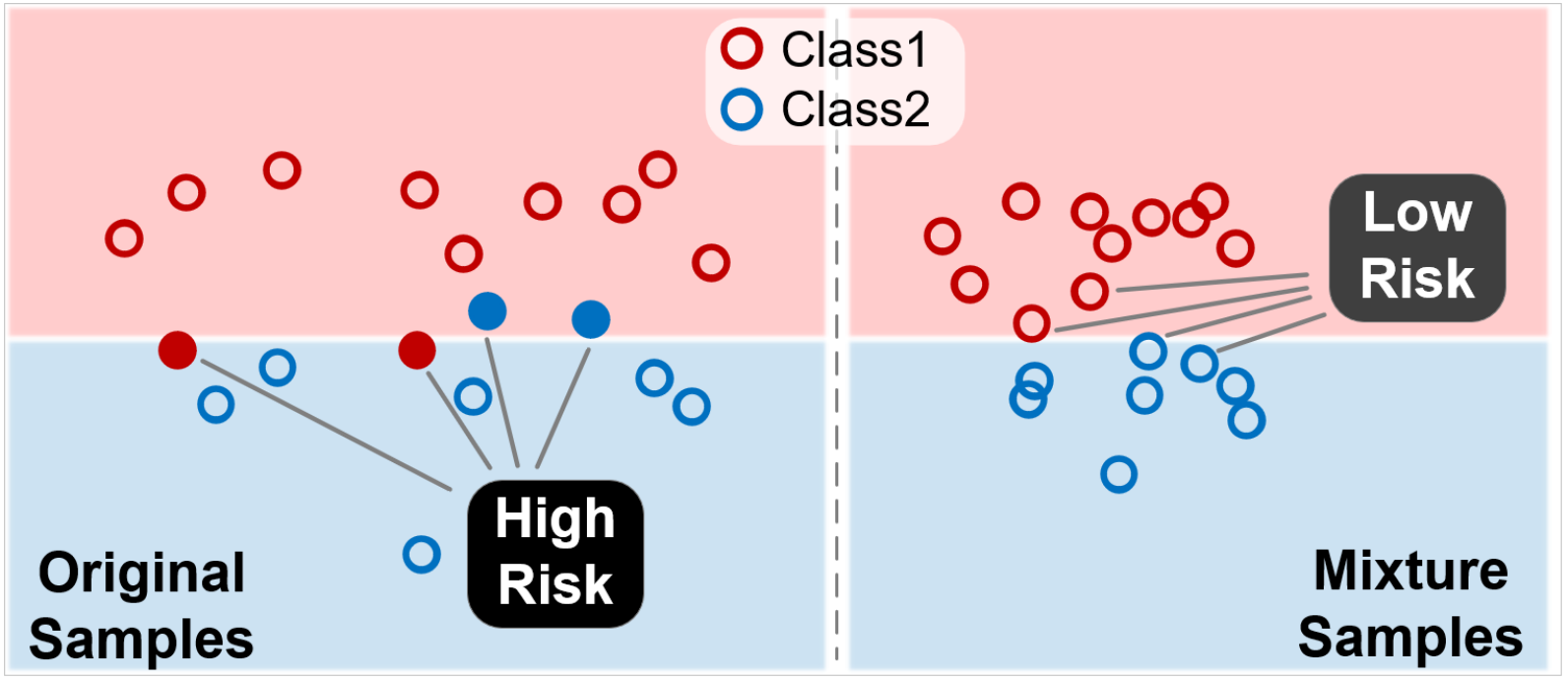} 
    \vspace{-16pt}
    \caption{A simple illustration of the reason for choosing $\theta\in(0,1)$. The LHS are the original samples corresponding to mix-ratio $\theta=0\; {\rm or}\;1$, while the RHS are the samples with mixture operation corresponding to $\theta\in(0,1)$. It is clear that the latter can significantly decrease the model risk.}
    \label{fig:fig1}
\end{figure}

We now give a more detailed explanation to understand better the proposed IS$^2$C. 

IS$^2$C forms a sampling domain by drawing labeled data $ \{\bm{x}_i^c,y_i^c\}_{i=1}^{n_c} $ from a sampling distribution $ P_c $ satisfies $ P_c(Y)=P_t(Y) $. When the sampling domain data $ \{\bm{x}_i^c,y_i^c\}_{i=1}^{n_c} $ are obtained, IS$^2$C trains the model by minimizing the empirical classification risk on $ P_c(X)=\frac{1}{n_c}\sum_{i=1}^{n_c} \delta_{\bm{x}_i^c}$ and explicitly aligning the class-conditional distribution. In contrast to many PDA methods whose learning procedures are inevitably affected by the outliers in practical applications, IS$^2$C trains with the new sampling data where the proportion of outliers is small, which significantly eliminates most of the harmful effects of outliers. Moreover, for any class $k$, samples $\{\bm{x}_i^c\}$ belonging to this class will be closer to the class prototype as the variance of class-conditional distribution decreases~\cite{carratino2020mixup}, thus it makes the cluster structure more evident. These advantages greatly avoid the harmful impact of outliers and ensure better model generalization ability.

\subsection{Generalization Error Analysis}

In this section, we focus on the reasoning of the proposed methodology from a theoretical view, where the performance guarantee for IS$^2$C is provided by proving the upper bound of the error gap between the sampling domain $P_c$ and the target domain $P_t$. The detailed proofs for theoretical results are provided in the Supplementary Material.

Before presenting the main results, we first introduce two important concepts for theoretical analysis as Combes et al.~\cite{tachet2020domain}. Given that the definitions in~\cite{tachet2020domain} are formulated under the assumption that the source and target label distributions share the same supports, they are not applicable for handling PDA tasks. Consequently, there is a necessity to reconstruct the theoretical underpinnings for PDA.

\begin{definition}[Balanced Prediction Error]
    The\;balanced prediction error of pseudo-label $\hat{Y}$ on distribution $P$ is
    \begin{equation*}\label{eq1}
        \Delta_{{\rm BE}}(P)(\hat{Y}\|Y) := \max \limits_{j\in \mathcal{Y}} p(\hat{Y}\neq j|Y=j).
    \end{equation*}
\end{definition}
\begin{definition}[Conditional Error Gap]
    Given the source distribution $ P_s $ and the target distribution $ P_t $, the conditional error gap of predictor $ \hat{Y} $ is
    \begin{align*}\label{eq2}
        \begin{split}
            \Delta_{{\rm CE}}(\hat{Y}):=\sum_{j=1}^{K}p_t(Y=j)\mathbb{D}(P_s(\hat{Y}|Y=j), P_t(\hat{Y}|Y=j)),
        \end{split}
    \end{align*}
    where
    \begin{align*}
        \mathbb{D}(P_s(\hat{Y}|Y=j & ), P_t(\hat{Y}|Y=j)) :=\\
        &\max_{i\neq j}|p_s(\hat{Y}=i|Y=j)-p_t(\hat{Y}=i|Y=j)|.
    \end{align*}

\end{definition}

Intuitively, balanced prediction error implies the maximum disagreement between true label $Y$ and pseudo-label $\hat{Y}$ and can be regarded as the degree of error for the prediction, and the risk minimization on $P$ is sufficient to minimize it~\cite{tachet2020domain}. The conditional error gap implies the cross-domain conditional discrepancy. Moreover, when  $ P_s(g(X)|Y=j)=P_t(g(X)|Y=j) $ for any $j\in {\rm supp}(P_t(Y))$, and $ \hat{Y} $ depends only on $ X $, the conditional error gap $\Delta_{{\rm CE}}(\hat{Y})$ will be zero.

Based on these two concepts, we present the main result for the generalization error of IS$^2$C as the following theorem.
\begin{theorem}\label{th1}
    Consider $X_1,X_2\sim P_s(X|Y = j)$ and $\theta \in[0,1]$. Then for a given distribution $P_c$ satisfies $ X=\theta X_1+(1-\theta)X_2 $ for variable $X\sim P_c(X|Y = j)$, $\forall j\in [K]$. If the model $ h\circ g $ is $\ell$-Lipschitz, then
    \begin{equation*}\label{eq3}
        \begin{aligned}
            |\varepsilon_t(h\circ g)&-\varepsilon_{c}(h\circ g)|\leq\left\|P_{c}(Y)-P_{t}(Y)\right\|_1\cdot \Delta_{{\rm BE}}(P_c)(\hat{Y}\|Y) \\&+ (K-1) \Delta_{{\rm CE}}(\hat{Y})+2\ell(K-1)\sqrt{\theta(1-\theta)}C_{s},
        \end{aligned}
    \end{equation*}
    where
    \begin{align*}
        \|P_{c}(Y)-P_{t}(Y)\|_1=&\sum_{i=1}^{K}\left|p_c(Y=i)-p_t(Y=i)\right|,\\
        C_{s}=&\max \limits_{j\in \mathcal{Y}} \mathbb{E}_{P_s (X|Y=j)}(\|X\|).
    \end{align*}
\end{theorem}

The upper bound in Theorem \ref{th1} provides an effective way to decompose the error gap between sampling and target domains, and it can also be directly used to obtain a generalization upper bound of the target risk $ \varepsilon_t $, i.e.,
\begin{equation}\label{eq6}
    \begin{aligned}
        \varepsilon_t&\leq\varepsilon_{c}+\left\|P_{c}(Y)-P_{t}(Y)\right\|_1\cdot \Delta_{{\rm BE}}(P_c)(\hat{Y}\|Y) \\&+ (K-1) \Delta_{{\rm CE}}(\hat{Y})+2\ell(K-1)\sqrt{\theta(1-\theta)}C_{s}.
    \end{aligned}
\end{equation}

This generalization upper bound, i.e., the RHS of \eqref{eq6}, consists of four terms. The first term is the risk on $P_c$, and IS$^2$C reduces it by minimizing the empirical risk on $P_c$.

\begin{figure*}[t] \centering
    \subfloat[Source]{
        \includegraphics[width=0.223\textwidth, height=0.175\textheight]{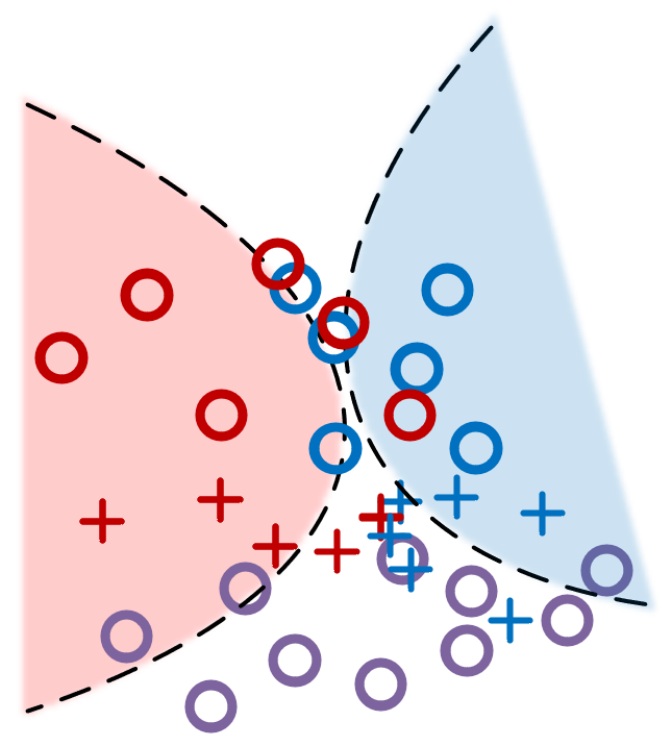}
        \label{fig:th1_Source}}
    \subfloat[$ \theta= 0$ or $1$]{
        \includegraphics[width=0.223\textwidth, height=0.175\textheight]{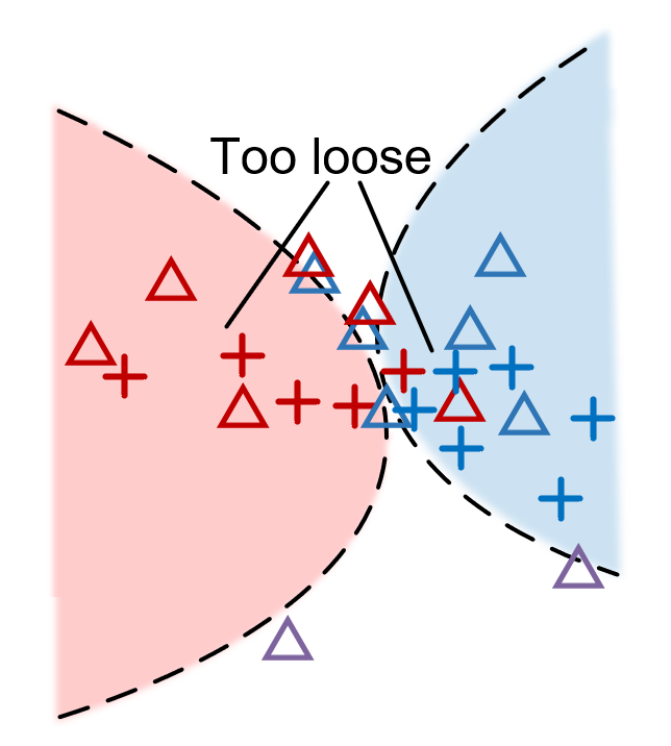}
        \label{fig:th1_theta_0or1}}
    \subfloat[$ \theta=0.5$]{
        \includegraphics[width=0.223\textwidth, height=0.175\textheight]{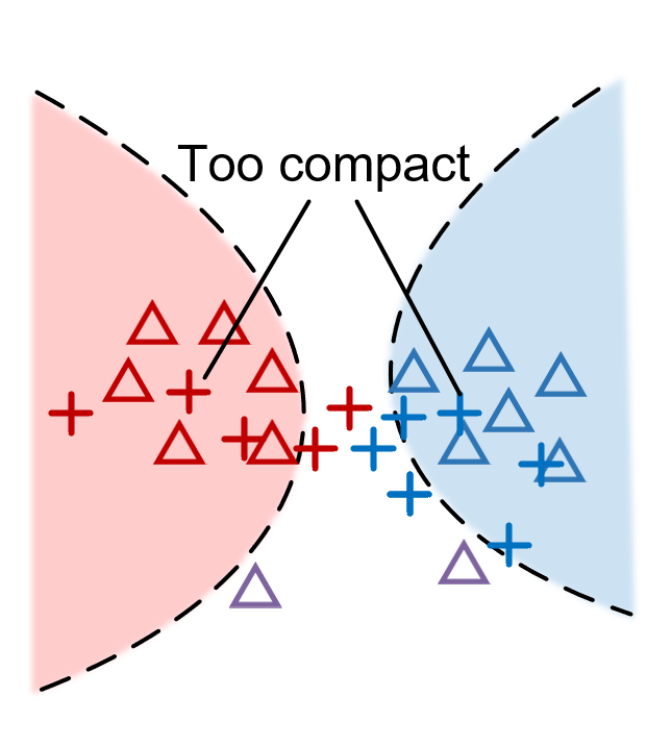}
        \label{fig:th1_theta_0point5}}
    \subfloat[Appropriate $\theta$]{
        \includegraphics[width=0.223\textwidth, height=0.175\textheight]{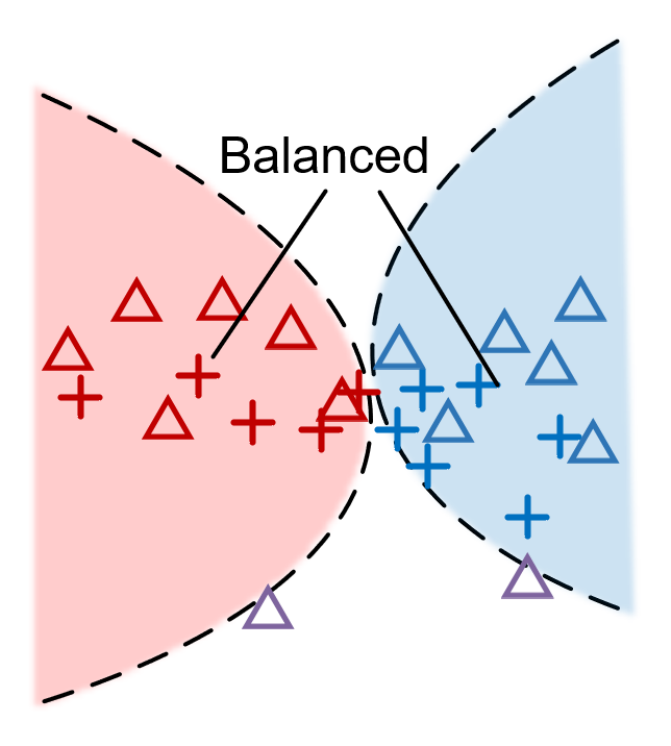}
        \label{fig:th1_optimal_theta}}
    \caption{An intuitive illustration of IS$^2$C under Theorem \ref{th1}. Different colors represent different classes.  ``$\circ$'' represents the source domain, ``$\vartriangle$'' represents the sampling domain, and ``+'' represents the target domain. The dashed lines represent decision boundaries. (a). The original source domain where exists label shift and conditional shift, and the target domain. (b)-(c). The sampling domains with different mix-ratio and the target domain. Though they achieve label correction and conditional correction, the cluster structure is too loose when $\theta = 0$ or $1$ while too compact when $\theta = 0.5$. (d). The sampling domain with an appropriate mix-ratio and the target domain.}
    \label{fig:th1_illustration}
\end{figure*}

The second term depends on the balanced prediction error on $P_{c}$ and $\|P_{c}(Y)-P_{t}(Y)\|_1$, which measures the distance between the label distributions across domains, and it can be large since there is no constraint on label distribution in Theorem \ref{th1}. However, IS$^2$C finds a sampling domain with distribution $P_{c}$ such that $ P_c(Y)$ closely approximates $P_t(Y) $ and implements risk minimization on $P_c$. This simultaneously reduces $\|P_{c}(Y)-P_{t}(Y)\|_1$ and the balanced prediction error, sufficiently minimizing the second term.  

The third term, by definition, measures the class-conditional distribution discrepancy on shared classes, i.e., classes in ${\rm supp}\left(P_t(Y)\right)$, between source and target domains. Most weighting-based methods minimize this term by aligning $P_s^{\bm{w}}(X)$ with $P_t(X)$, which can be seen as implicitly aligning $P_s(g(X)|Y)$ with $P_t(g(X)|Y)$ \cite{tachet2020domain} under some assumptions. However, the assumptions may be too strict to satisfy in practical applications and such alignment tends to cause harmful misalignment. By contrast, IS$^2$C prevents harmful misalignment by explicitly aligning $P_s(g(X)|Y)$ with $P_t(g(X)|Y)$ on shared classes to minimize the third term.

The fourth term is a constant no less than zero given mix-ratio $\theta$ and source domain. In practice, what we can change is $\theta$. It is obvious that the fourth term is equal to 0 when $\theta=0$ or $1$. This corresponds to directly selecting source data as sampling domain data in the sampling procedure. However, as shown in Figure~\ref{fig:fig1}, we still choose $\theta\in (0,1)$ by the intuition that this would help reduce the first term; moreover, the relation between $\theta$ and training risk $\varepsilon_{c}$ can be explicitly analyzed under some mild assumption.

To provide a better understanding on the risk of the model learned on sampling source, we next present an important proposition to show the advantage of the importance sampling strategy. Under the convexity assumption, the next result shows that the sampling source domain generally ensures a smaller risk compared with original source domain.
\begin{prop}\label{pr1}
    Given $P_c$ and $P_{c'}$, defined as in Theorem \ref{th1}, where $P_c$ corresponds to $\theta\in(0,1)$ and $P_{c'}$ corresponds to $\theta=0$ or $1$ respectively. If $P_c$ and $P_{c'}$ have the same label distribution and each component of the model output $ h\circ g(\cdot)_i $ is a convex function of the input, then
    \begin{equation*}
            \varepsilon_c(h\circ g)\leq\varepsilon_{c'}(h\circ g),
    \end{equation*}
    where the equality holds when $h\circ g$ is linear model.
\end{prop}

Note that for the advanced methods, e.g., neural network methods and kernel methods, the model is usually non-linear. Therefore, the strict smaller risk of sampling source generally holds, i.e., $\varepsilon_c(h\circ g)<\varepsilon_{c'}(h\circ g)$. So there is a trade-off between the increase of the fourth term and the decrease of the first term. Since the minimization of classification loss is equivalent to risk minimization on reweighed source domain (coined as wERM, corresponding to the IW strategy) when $\theta=0$ or $1$, our experiments show that using a mixture conditional distribution, i.e., choosing $\theta\in (0,1)$, is indeed an efficient way to improve the model performance. 


In summary, for an IS$^2$C learning model, the second and third terms in the generalization upper bound are expected to approach zero, while the first and fourth terms will be balanced to minimize their sum. Therefore, an IS$^2$C learning model can achieve a smaller generalization error on the target domain. An intuitive illustration of IS$^2$C under Theorem \ref{th1} is presented in Figure \ref{fig:th1_illustration}. As shown in Figure \ref{fig:th1_Source}, there exists label shift and conditional shift in the original source. Through IS$^2$C, the sampling domains achieve label correction and conditional correction as shown in Figures \ref{fig:th1_theta_0or1}-\ref{fig:th1_optimal_theta}. However, the cluster structure is too loose when $\theta=0$ or $1$ while too compact when $\theta=0.5$ as shown in Figures \ref{fig:th1_theta_0or1}-\ref{fig:th1_theta_0point5}, and with an appropriate $\theta$ it reaches balanced, which corresponds to the trade-off as aforementioned.

\section{IS$^2$C: the  algorithm}\label{sec4}

Now, we introduce a specific learning algorithm under the IS$^2$C. First, we focus on the way to align the class-conditional distributions across domains.

\subsection{ETIC-based Alignment}

We derive an alignment loss based on the entropy-regularized optimal transport independence criterion (ETIC) \cite{liu2022entropy}. Let $\mathcal{Z}=\{\bm{z}^s,\bm{z}^t\}$ be a domain label space, where $\bm{z}^s=[0\;1]$ is a $2$-dimensional vector representing the source domain and $\bm{z}^t=[1\;0]$ representing the target domain. We use $Z$ to denote random variable which takes values in $\mathcal{Z}$ and consider the distribution $ P(g(X),Y,Z) $ on $\mathcal{F}\times \mathcal{Y} \times \mathcal{Z}$, where:
\begin{align*}
    P(g(X)|Y,Z=\bm{z}^s)&=P_s(g(X),Y),\\
    P(g(X)|Y,Z=\bm{z}^t)&=P_t(g(X),Y),
\end{align*}
If $Z$ and $g(X)$ are conditionally independent given $Y$, then $P(g(X)|Y,Z)=P(g(X)|Y)$, i.e., $P_s(g(X),Y)=P(g(X)|Y,Z=\bm{z}^s)=P(g(X)|Y,Z=\bm{z}^t)=P_t(g(X),Y)$. Therefore, a way to achieve alignment is training the feature transformation $g(\cdot)$ to maximize the independence between $Z$ and $g(X)$ given $Y$.

To measure the independence between two random variables, we introduce ETIC instead of the commonly used Hilbert-Schmidt independence criterion (HSIC) \cite{gretton2005kernel,gretton2007kernel} since the high interest of ETIC to capture both linear and nonlinear dependencies in synthetic data and real data \cite{liu2022entropy}.
ETIC takes the symmetric sinkhorn divergence \cite{feydy2019interpolating} between the product distribution and the joint distribution as the independence measure. For convenience, we use $P_{XZ}^j$, $P_{X}^j$, $P_{Z}^j$ to respectively represent $P(g(X),Z|Y=j)$, $P(g(X)|Y=j)$, $P(Z|Y=j)$, then ETIC is defined as
\begin{equation*}
    \begin{aligned}
        T_{\epsilon}(g(X),Z|Y=j) :=&\ S_{\epsilon}(P_{XZ}^j, P_{X}^j \otimes P_{Z}^j) \\
        &\ -\frac{1}{2}S_{\epsilon}(P_{XZ}^j, P_{XZ}^j) \\
        &\ -\frac{1}{2}S_{\epsilon}(P_{X}^j \otimes P_{Z}^j, P_{X}^j \otimes P_{Z}^j).
    \end{aligned}
\end{equation*}
Here $S_{\epsilon}(P_{XZ}^j,P_{X}^j\otimes P_{Z}^j)$ is the entropy-regularized optimal transport cost between $P_{XZ}^j$ and $P_{X}^j\otimes P_{Z}^j$, i.e.,
\begin{equation*}\label{eq10}
    \begin{aligned}
        \inf_{\gamma}\left[\int c d\gamma+\epsilon {\rm KL}\left(\gamma\|P_{XZ}^j\otimes\left(P_{X}^j\otimes P_{Z}^j\right)\right) \right]
    \end{aligned}
\end{equation*}
where the infimum is over $\Pi(P_{XZ}^j,P_{X}^j\otimes P_{Z}^j)$ which is the set of couplings on $\mathcal{F}\times \mathcal{Z}$ with marginals $P_{XZ}^j$ and $P_{X}^j\otimes P_{Z}^j$, $\epsilon > 0$ is the regularization coefficient, and KL is the Kullback-Leibler divergence. The other two terms are defined similarly and are omitted for the sake of space. $c$ is the addictive cost functions, i.e., $c((g(\bm{x}),\bm{z}),(g(\bm{x}'),\bm{z}'))=c_1(g(\bm{x}),g(\bm{x}'))+c_2(\bm{z},\bm{z}')$, $\forall (g(\bm{x}),\bm{z}),(g(\bm{x}'),\bm{z}') \in \mathcal{F}\times \mathcal{Z}$, where both $c_1$ and $c_2$ are Euclidean distance. Following \cite{liu2022entropy}, for this type of cost functions, we have $T_{\epsilon}(g(X),Z|Y)\geq 0$ and
\begin{align*}
    T_{\epsilon}(g(X),Z|Y=j) =0\;{\rm iff}\;P_{XZ}^j=P_{X}^j\otimes P_{Z}^j,
\end{align*}
i.e., $P_s(g(X)|Y=j) =P_t(g(X)|Y=j)$ if and only if $T_{\epsilon}(g(X),Z|Y=j) =0$. Hence the alignment loss term can be defined as
\begin{align*}
    T_{\epsilon}(g(X),Z):=\sum_{j=1}^{K} p_t(Y=j)T_{\epsilon}(g(X),Z|Y=j).
\end{align*}
It is obvious that we can achieve class-conditional alignment on shared classes by training the feature transformation $g(\cdot)$ to minimize $T_{\epsilon}(g(X),Z)$.

In practice, as aforementioned, the distributions are denoted as empirical distributions. Therefore, for each class $j$, we can only compute the empirical estimate of $T_{\epsilon}(g(X),Z|Y=j)$. The Tensor Sinkhorn algorithm \cite{liu2022entropy} has been proposed to obtain the estimation. However, directly using the algorithm for PDA without adjustment takes $\mathcal{O}(n_j^3)$ time each iteration, where $n_j$ is the number of source and target domain samples belonging to class $j$. We will show that in practical PDA applications, the computation can indeed be sped up.

\subsection{An Adjusted Computation for ETIC}

The original Tensor Sinkhorn algorithm considers the prior assumption that both $X$ and $Z$ are continuous variables. Therefore, the support of empirical distribution for $P_{XZ}^j$ or $P_{X}^j\otimes P_{Z}^j$ contains at most $n_j^2$ points when the number of samples belonging to class $j$ is $n_j$. As a result, the matrices used to represent $P_{XZ}^j$, $P_{X}^j\otimes P_{Z}^j$, and the transport cost must be square matrices of order $n_j$, leading to high complexity in matrix computations. However, in practical PDA applications, the domain variable $Z$ is a discrete variable with binary values. So, each of the supports contains at most $2n_j$ points, and the matrix computations during the iteration process can be accelerated with smaller matrix sizes utilizing this prior knowledge. We next illustrate the computation algorithm on the estimation of $S_{\epsilon}(P_{XZ}^j,P_{X}^j\otimes P_{Z}^j)$ as example.

Put together the source and target domain features that belong to class $j$, denoted as $\{g(\bm{x}_i),\bm{z}_i\}_{i=1}^{n_j}$, where $\bm{z}_i\in \mathcal{Z}$ is the domain label corresponds to $\bm{x}_i$, and $\bm{z}_i=\bm{z}^s$ for $i=1,\cdots,n_j^s$, while the remaining $n_j^t=n_j-n_j^s$ are equal to $\bm{z}^t$. Since there are no labels for target domain samples, we use the pseudo label $\hat{y}=\argmax_{i}(h\circ g(\bm{x}))_i$, where $(h\circ g(\bm{x}))_i$ is the $i$-th component of $h\circ g(\bm{x})$. The empirical distributions can be written as
\begin{eqnarray*}
    P_{XZ}^j \!\!&=&\!\!\frac{1}{n_j}\sum_{i=1}^{n_j} \delta_{(g(\bm{x}_i),\bm{z}_i)},\\
     P_{X}^j \!\!&=&\!\! \frac{1}{n_j}\sum_{i=1}^{n_j} \delta_{g(\bm{x}_i)},\\
     P_{Z}^j \!\!&=&\!\! \frac{n_j^s}{n_j}\delta_{\bm{z}^s}+\frac{n_j^t}{n_j}\delta_{\bm{z}^t}.
\end{eqnarray*}

Let matrices $\bm{A}$ and $\bm{B}\in \mathbb{R}^{n_j \times 2}$ represent $P_{XZ}^j$ and $P_{X}^j\otimes P_{Z}^j$ respectively, where the $i$-th row of $\bm{A}$ is $\bm{A}_i=\bm{z}_i/n_j$ and the $i$-th row of $\bm{B}$ is $\bm{B}_i=[\frac{n_j^s}{n_j}\;\frac{n_j^t}{n_j}]$. Consider the mentioned addictive cost function $c=c_1+c_2$. Let $\bm{C}_1\in \mathbb{R}^{n_j \times n_j}$ be the cost matrix of $\{g(\bm{x}_i)\}_{i=1}^{n_j}$ and $\bm{C}_2 \in \mathbb{R}^{2 \times 2}$ be the cost matrix of $\{\bm{z}^s,\bm{z}^t\}$. Define kernel matrices $\bm{K}_1:=\exp(-\bm{C}_1/(\lambda_1\epsilon))$ and $\bm{K}_2 := \exp(-\bm{C}_2/(\lambda_2\epsilon))$, where the exponential function is element-wise and $\lambda_1, \lambda_2 > 0$ are kernel parameters. Then we can choose initial matrices $\bm{U}^{(0)}$,$\bm{V}^{(0)} \in \mathbb{R}^{n_j \times 2}$ and use the fixed point iterations similar to Tensor Sinkhorn algorithm:
\begin{eqnarray}\label{sinkhorn iteration}
  \nonumber \bm{U}^{(i)} \!\!&=&\!\! \bm{A}\oslash(\bm{K}_1\bm{V}^{(i-1)}\bm{K}_2^T), \\
  \bm{V}^{(i)} \!\!&=&\!\! \bm{B}\oslash(\bm{K}_1\bm{U}^{(i)}\bm{K}_2^T) ,
\end{eqnarray}
where $\oslash$ represents element-wise division. Now we can analyze the computation cost of our iteration algorithm. Since $\bm{K}_1\in \mathbb{R}^{n_j \times n_j}$,$\bm{V}^{(i-1)}\in \mathbb{R}^{n_j \times 2}$ and $\bm{K}_2\in \mathbb{R}^{2 \times 2}$, the cost of computing $\bm{K}_1\bm{V}^{(i-1)}\bm{K}_2^T$ is $n_j\cdot n_j\cdot2+n_j\cdot2\cdot2=2n_j^2+4n_j$ by the rules of matrix multiplication. Meanwhile, both $\bm{K}_1\bm{V}^{(i-1)}\bm{K}_2^T$ and $\bm{A}$ belong to $\mathbb{R}^{n_j \times 2}$, so the division in $\bm{A}\oslash(\bm{K}_1\bm{V}^{(i-1)}\bm{K}_2^T)$ takes $2n_j$ time and thus the total computation cost of $\bm{A}\oslash(\bm{K}_1\bm{V}^{(i-1)}\bm{K}_2^T)$ is $2n_j+2n_j^2+4n_j=2n_j^2+6n_j$. Similarly, we can derive that $\bm{B}\oslash(\bm{K}_1\bm{U}^{(i)}\bm{K}_2^T)$ takes the same total cost of $2n_j^2+6n_j$ and finally our algorithm takes $4n_j^2+12n_j=\mathcal{O}(n_j^2)$ time each iteration. For comparison, we further present the computation cost of the Tensor Sinkhorn algorithm. During the iteration process of the Tensor Sinkhorn algorithm, all matrices are square matrices of order $n_j$. Hence the computation cost is $4n_j^3+2n_j^2=\mathcal{O}(n_j^3)$ each iteration. Compared with the original Tensor Sinkhorn algorithm, our variant sufficiently reduces the computation cost from $\mathcal{O}(n_j^3)$ to $\mathcal{O}(n_j^2)$ each iteration.

After $L$ iterations, we get $\bm{U}=\bm{U}^{(L)}$, $\bm{V}=\bm{V}^{(L)}$, and then $S_{\epsilon}(P_{XZ}^j,P_{X}^j\otimes P_{Z}^j)$ can be approximated by
\begin{align}
    \begin{split}\label{eq:sinkhorn estimation}
        \hat{S}_{\epsilon}(P_{XZ}^j,P_{X}^j\otimes P_{Z}^j) :=\ &{\rm sum}\Big(\bm{U}\odot\left[ \bm{K}_1\bm{V}(\bm{K}_2\odot \bm{C}_2)^T \right] \\
        &+ \bm{U}\odot\left[(\bm{K}_1\odot \bm{C}_1)\bm{V}\bm{K}_2^T \right] \Big),
    \end{split}
\end{align}
where ${\rm sum}(\bm{A})=\sum_{i,j}\bm{A}_{i,j}$, and $\odot$ represents element-wise multiplication. The empirical estimate of $\hat{S}_{\epsilon}(P_{XZ}^j,P_{XZ}^j)$, $\hat{S}_{\epsilon}(P_{X}^j\otimes P_{Z}^j,P_{X}^j\otimes P_{Z}^j)$ can be similarly computed by replacing the matrix $\bm{A}$ or $\bm{B}$ while the other steps remain the same.

Finally, our ETIC-based alignment loss for practical training is formulated as
\begin{align}\label{ali_loss}
    \mathcal{L}_{align}:=\sum_{j=1}^{K} p_t(Y=j)\hat{T}_{\epsilon}(g(X),Z|Y=j),
\end{align}
where
\begin{equation}\label{eq:ETIC estimation}
    \begin{aligned}
        \hat{T}_{\epsilon}(g(X),Z|Y=j) :=&\ \hat{S}_{\epsilon}(P_{XZ}^j, P_{X}^j \otimes P_{Z}^j) \\
        &\ -\frac{1}{2}\hat{S}_{\epsilon}(P_{XZ}^j, P_{XZ}^j) \\
        &\ -\frac{1}{2}\hat{S}_{\epsilon}(P_{X}^j \otimes P_{Z}^j, P_{X}^j \otimes P_{Z}^j).
    \end{aligned}
\end{equation}

\begin{algorithm}[!t]
    \caption{Partial domain adaptation via IS$^2$C}
    \begin{algorithmic}[1]\label{Ag2}
        \renewcommand{\algorithmicrequire}{\textbf{Input:}}
        \renewcommand{\algorithmicensure}{\textbf{Output:}}
        \REQUIRE source domain data $\{\bm{x}_i^s,y_i^s\}_{i=1}^{n^s}$, target domain data $\{\bm{x}_i^t\}_{i=1}^{n^t}$, parameters $\mu$, $\lambda_1$, $\lambda_2$, $\theta$, $n_c$, $\epsilon$, $T$;
        \ENSURE  $h(\cdot)$ and $g(\cdot)$;\\
        \STATE Approximate $p_s(Y)$ by source data $\{\bm{x}_i^s,y_i^s\}_{i=1}^{n^s}$;
        \FOR {$i = 1$ to $T$}
        \STATE Using the model $h\circ g$ to give each target sample a pseudo label;
        \STATE Estimate $\dfrac{p_t(Y)}{p_s(Y)}$ with BBSE algorithm and get the estimation of $p_t(Y)$ by $p_t(Y)=p_s(Y)\dfrac{p_t(Y)}{p_s(Y)}$;
        \STATE Execute the importance sampling process and obtain $\{\bm{x}_i^c,y_i^c\}_{i=1}^{n^c}$;
        \STATE Compute the empirical risk by Eq.~\eqref{risk};
        \STATE Compute the alignment loss by Eqs.~\eqref{eq:sinkhorn estimation}-\eqref{eq:ETIC estimation}, and the iteration formula \eqref{sinkhorn iteration};
        \STATE Update $h(\cdot)$ and $g(\cdot)$ by Eq.~\eqref{final_obj}.
        \ENDFOR
    \end{algorithmic}
\end{algorithm}

\begin{table*}[t]
    \renewcommand{\arraystretch}{0.8}
    \renewcommand{\tabcolsep}{0.25pc} 
    \begin{center}
        \caption{Classification accuracies ($ \% $) on Office-Home and VisDA-2017 datasets (ResNet-50). }
        \label{tab:Home_VisDA}
        \begin{tabular}{c*{13}{c}c}
            \toprule
            \multirow{2}{*}{\textbf{Methods}} & \multicolumn{13}{c}{\textbf{Office-Home}} &
            \textbf{VisDA-2017}\\
            & Ar$ \to $Cl	& Ar$ \to $Pr	&Ar$ \to $Rw	&Cl$ \to $Ar	&Cl$ \to $Pr	&Cl$ \to $Rw	&Pr$ \to $Ar	&Pr$ \to $Cl	&Pr$ \to $Rw	&Rw$ \to $Ar	&Rw$ \to $Cl	&Rw$ \to $Pr& \textbf{Mean}	&S$ \to $R\\
            \midrule
            Baseline \cite{he2016deep}&46.3&67.5&75.9&59.1&59.9&62.7&58.2&41.8&74.9&67.4&48.2&74.2&61.4&45.3\\
            DANN \cite{ganin2016domain}&43.8&67.9&77.5&63.7&59.0&67.6&56.8&37.1&76.4&69.2&44.3&77.5&61.7&51.0\\
            PADA \cite{cao2018partial}&52.0&67.0&78.7&52.2&53.8&59.0&52.6&43.2&78.8&73.7&56.6&77.1&62.1&53.5\\
            ETN \cite{cao2019learning}&59.2&77.0&79.5&62.9&65.7&75.0&68.3&55.4&84.4&75.7&57.7&84.5&70.5&-\\
            HAFN \cite{xu2019larger}&53.4&72.7&80.8&64.2&65.3&71.1&66.1&51.6&78.3&72.5&55.3&79.0&67.5&65.1\\
            SAFN \cite{xu2019larger}&58.9&76.3&81.4&70.4&73.0&77.8&72.4&55.3&80.4&75.8&60.4&79.9&71.8&67.7\\
            DMP \cite{luo2020unsupervised}&59.0 &81.2 &86.3 &68.1 &72.8 &78.8 &71.2 &57.6 &84.9 &77.3 &61.5 &82.9 &73.5 &72.7\\
            AR \cite{gu2021adversarial}& \textbf{67.4} &85.3 &90.0 & 77.3 &70.6 &85.2 &79.0 & \textbf{64.8} &89.5 &80.4 &66.2 &86.4 &78.3 &88.7\\
            m-POT \cite{nguyen2022improving}&64.6 &80.6 &87.2&76.4&77.6&83.6&77.1&63.7&87.6&81.4& \textbf{68.5}&87.4&78.0&87.0\\
            IDSP~\cite{li2022idsp}&60.8&80.8&87.3&69.3&76.0&80.2&74.7&59.2&85.3&77.8&61.3&85.7&74.9&-\\            SAN++~\cite{cao2022san}&61.3&81.6&88.6&72.8&76.4&81.9&74.5&57.7&87.2&79.7&63.8&86.1&76.0&63.1\\
            RAN~\cite{wu2022ran}&63.3&83.1&89.0&75.0&74.5&82.9&78.0&61.2&86.7&79.9&63.5&85.0&76.8&75.1\\
            MUL \cite{luo2022generalized}&57.4 & \textbf{88.7} &90.8 &71.0 &80.4 &82.1 &77.9 &59.8 &\textbf{91.2} & 83.5 &58.1 & 87.7 &77.4 &77.5\\
            CI~\cite{lin2022CI}&61.7&86.9&90.5&77.2&76.9&83.8&79.6&63.8&88.5&85.0&65.8&86.2&78.8&69.8\\
            SLM~\cite{sahoo2023select}&61.1&84.0&91.4&76.5&75.0&81.8&74.6&55.6&87.8&82.3&57.8&83.5&76.0& 91.7\\
            MOT~\cite{luo2023mot} &63.1 &86.1 &\textbf{92.3} &\textbf{78.7} &\textbf{85.4} &\textbf{89.6} &\textbf{79.8} &62.3 &89.7 &\textbf{83.8} &67.0 &\textbf{89.6} &\textbf{80.6} &\textbf{92.4}\\
            
            \midrule
            IS$^2$C& 61.1	&87.9	& \textbf{91.5}	&74.4	& \textbf{81.0}	& \textbf{87.2}	& \textbf{80.7}	&60.5	& \textbf{91.7}	&81.7	&65.1	&87.4	& \textbf{79.2}	& {89.3}\\
            \bottomrule
        \end{tabular}
    \end{center}
\end{table*}

\subsection{Training Algorithm}

In this section, we present the overall training algorithm for IS$^2$C, which is also summarized in Algorithm \ref{Ag2}.

In each iteration, we first need to obtain the new sampling domain data $\{\bm{x}_i^s,y_i^s\}_{i=1}^{n_s}$. Recall the IS$^2$C in definition \ref{Def_I2SC}, we practically perform the sampling process by following steps:
\begin{itemize}
    \item Sample a $ j\in \mathcal{Y}$ according to the class proportions $p_t(Y)$.
    \item Randomly select two source samples $\bm{x}_k^s$ and $\bm{x}_l^s$ from $ \{\bm{x}_i^s\}_{i=1}^{n_s} $ where $y_k^s=y_l^s=j$. Then the labeled sampling domain data from $P_c$ can be obtained by $\bm{x}^c= \theta \bm{x}_k^s+ (1-\theta) \bm{x}_l^s $ and $y^c=j$, where $\theta$ is the mix-ratio.
    \item Repeat the above steps until there are $n_c$ labeled sampling domain data.
\end{itemize}

Since we have no prior knowledge of $p_t(Y)$, we estimate $\dfrac{p_t(Y)}{p_s(Y)}$ with the BBSE algorithm~\cite{lipton2018detecting}, and approximate $p_t(Y)$ by $p_t(Y)=p_s(Y)\dfrac{p_t(Y)}{p_s(Y)}$. To employ BBSE, we rewrite $\dfrac{p_t(Y)}{p_s(Y)}$ and $p_s(Y)$ as $K$-dimensional vectors $\bm{a}$ and $\bm{p}_Y^s$, respectively. The $j$-th component of $\bm{p}_Y^s$ equal to $n_j^s/n^s$. Similarly, based on the predictor’s output, the
plug-in estimation~\cite{lipton2018detecting} of $p_t(\hat{Y})$ and $p_s(\hat{Y},Y)$ are rewritten as $\bm{p}_{\hat{Y}}^t\in \mathbb{R}^{K}$ and $\bm{M}\in \mathbb{R}^{K\times K}$, respectively. Then the BBSE algorithm can be implemented by solving a Quadratic-Programming problem~\cite{tachet2020domain}:
\begin{align*}
\min_{\bm{a}}\|\bm{p}_{\hat{Y}}^t-\bm{M}\bm{a}\|_{2}^{2}\quad\mathrm{~s.t.~}\bm{a}\geq0,\bm{a}^{T}\bm{p}_Y^s=1,
\end{align*}
where $\geq$ is element-wise operation. When the number of classes $K$ is not excessively large, the above Quadratic-Programming problem can be efficiently solved in time $\mathcal{O}(K^3)$~\cite{tachet2020domain}. The convergence property of BBSE is also theoretically ensured~\cite{lipton2018detecting}.


Secondly, we compute the empirical risk on $P_c$ and the alignment loss term. The alignment loss term $\mathcal{L}_{align}$ can be computed by Eq.~\eqref{ali_loss}, aiming to align the conditional feature distributions $P_s(g(X)|Y)$ and $P_t(g(X)|Y)$ as previously discussed. The empirical risk on $P_c$ is formulated as
\begin{align}\label{risk}
    \mathcal{L}_{risk}:= \frac{1}{n_c}\sum_{i=1}^{n_c} l(\hat{y}_i^c,y_i^c).
\end{align}
Here $l(\hat{y}_i^c,y_i^c)$ is the cross-entropy between $p_c(\hat{y}_i^c|\bm{x}_i^c)$ and $p_c(y_i^c|\bm{x}_i^c)$, where $p_c(\hat{y}_i^c|\bm{x}_i^c)=h\circ g(\bm{x}_i^c)$ and $p_c(y_i^c|\bm{x}_i^c)$ is a $K$-dimensional one-hot vector with its $y_i^c$-th component equal to $1$. Motivated by our theoretical analysis in Section~\ref{sec3}, this risk minimization on \( P_c \) is designed to reduce both the model error \( \varepsilon_c \) and the balanced prediction error \( \Delta_{\rm BE}(P_c) \). These two losses together serve to control the proposed upper bound.

Let $\mu>0$ be a parameter that balances the risk term and alignment term. Our final objective can be written as
\begin{align}\label{final_obj}
    \min_{g,h} \mathcal{L}=\mathcal{L}_{risk}+\mu \mathcal{L}_{align}.
\end{align}

We employ ResNet-50 pre-trained on ImageNet~\cite{deng2009imagenet} as backbone, which has been widely used in PDA methods, and fix it to extract 2048-dimensional features as input for all datasets. The transformation $ g(\cdot) $ consists of two fully connected layers, and the classifier $ h(\cdot) $ is a single fully connected layer with $ K $-dimensional output and SoftMax activation.

We initially fine-tune $ g(\cdot) $ and $ h(\cdot) $ with the empirical risk on source domain distribution $P_s$. Then we optimize the whole model $h\circ g$ by Algorithm \ref{Ag2}. The parameters of $ g(\cdot) $ and $ h(\cdot) $  are updated by the Adam optimizer~\cite{kingma2014adam} with learning rate 1$e$-3. The mix-ratio $\theta$ is randomly sampled from a beta distribution ${\rm Beta}(\alpha, \alpha)$, where sampling parameter $\alpha=0.2$. The sample size of the new sampling domain is set as $n_c=2n_s$. Following literature \cite{liu2022entropy}, we set the entropy regularization coefficient as $ \epsilon = 1 $ on all experiments, and kernel parameter $ \lambda_i $ of kernel matrix $ K_i $ as $\lambda_1=4m_x$ and $\lambda_2=1$, where $ m_x $ is the median of elements in cost matrix $ \bm{C}_1 $. More details about the training implementations are provided in the supplementary material.

\section{Numerical Experiments}\label{sec5}

\subsection{Datasets and Setup}

We evaluate IS$^2$C model on four PDA benchmark datasets, i.e., Image-CLEF \cite{caputo2014imageclef}, Office-31 \cite{saenko2010adapting},  Office-Home \cite{venkateswara2017deep} and VisDA-2017 \cite{peng2017visda}. To simulate PDA setting, we follow the common protocol~\cite{cao2018partial,luo2020unsupervised} to generate the target domain with a smaller label space. Specifically, the details for datasets and protocol are summarized as follows.

\begin{itemize}
    \item \textbf{Office-31} contains 4652 images of 31 categories. The images are collected from three different domains, including Amazon (A), DSLR (D), and Webcom (W). Following the literature \cite{cao2018partial}, 10 categories are selected as target categories, and 6 PDA tasks are built on 3 domains.
    \item \textbf{Image-CLEF} contains 3 domains, i.e., Caltech(C), ImageNet (I), and Pascal (P), that are collected from existing public datasets. Each domain contains 600 images of 12 categories. The first 6 categories (in alphabetical order) are selected as target categories, and 6 PDA tasks are built on 3 domains.
    \item \textbf{Office-Home} contains 15500 images of 65 categories from 4 domains: Art (Ar), Clipart (Cl), Product (Pr), and Real-World (Rw). The first 25 categories (in alphabetical order) are selected as target categories, and 12 PDA tasks are built on 4 domains.
    \item \textbf{VisDA-2017} consists of about 200k images of 12 categories from 2 domains: Synthetic (S) and Real (R). The first 6 classes (in alphabetical order) are selected as the target categories, then PDA task is considered as the knowledge transfer from S to R.
\end{itemize}

\subsection{Experimental results}

\begin{table}[!t]
    \renewcommand{\arraystretch}{0.8}
    \begin{center}
        \caption{Classification accuracies ($ \% $) on Office-31 dataset (ResNet-50). }
        \label{tab:Office31}
        \renewcommand{\tabcolsep}{0.25pc} 
        \begin{tabular}{c*{7}{c}}
            \toprule
            \textbf{Office-31}&A$\to$W&D$\to$W&W$\to$D &A$\to$D&D$\to$A&W$\to$A
            &\textbf{Mean}	\\
            \midrule
            Baseline \cite{he2016deep}&75.6&96.3&98.1&83.4&83.9&85.0&87.1\\
            DANN \cite{ganin2016domain}&73.6&96.3&98.7&81.5&82.8&86.1&86.5\\
            PADA \cite{cao2018partial}&86.5&99.3&\textbf{100}&82.2&92.7&95.4&92.7\\
            HAFN \cite{xu2019larger}&87.5&96.7&99.2&87.3&89.2&90.7&91.7\\
            SAFN \cite{xu2019larger}&87.5&96.6&99.4&89.8&92.6&92.7&93.1\\
            DMP \cite{luo2020unsupervised}&96.6 &\textbf{100} &\textbf{100} &96.4 &95.1 &95.4 &97.2 \\
            AR \cite{gu2021adversarial}&93.5 &\textbf{100} &99.7&96.8&95.5&96.0&96.9\\
            IDSP~\cite{li2022idsp}&99.7&99.7&\textbf{100}&99.4&95.1&95.7&98.3\\
            SAN++~\cite{cao2022san}&99.7&\textbf{100}&\textbf{100}&98.1&94.1&95.5&97.9\\
            RAN~\cite{wu2022ran}&99.0&\textbf{100}&\textbf{100}&97.7&96.3&96.2&98.2\\
            MUL \cite{luo2022generalized}&94.2& \textbf{100}& \textbf{100}& 98.5& 95.6& 96.3& 97.5\\
            CI~\cite{lin2022CI}&99.7&\textbf{100}&\textbf{100}&96.8&96.1&\textbf{96.6}&98.2\\
            SLM \cite{sahoo2023select}&99.8&\textbf{100}&99.8&98.7&96.1&95.9&98.4\\
            MOT~\cite{luo2023mot}&99.3 &\textbf{100} &\textbf{100} &98.7 &96.1 &96.4 &98.4\\
            \midrule
            IS$^2$C& \textbf{100}& \textbf{100}	& \textbf{100}	& \textbf{100}	& \textbf{96.5}	& 96.3	& \textbf{98.8}							
            \\
            \bottomrule
        \end{tabular}
    \end{center}
\end{table}

The results on Office-Home dataset are presented in Table~\ref{tab:Home_VisDA}. Compared with the Baseline model, the classical UDA method DANN cannot ensure significant performance improvement as it cannot deal with the outlier classes. By introducing label distribution-based reweighing and sample selection into adversarial training, PADA and ETN mitigate the impact of outlier classes and improve the accuracy significantly. Besides, compared with the IW-based methods, e.g., DMP, AR and MUL, the proposed IS$^2$C method has superior performance and the improvement on accuracy is about 1.0\% at least.

The results on VisDA-2017 dataset are also presented in Table~\ref{tab:Home_VisDA}. From the results, we can also observe that the label shift correction is indeed important to improve the performance on the target domain. Besides, compared with related PDA methods, the accuracy of IS$^2$C is only lower than the selection module-based method SLM, and ensures better performance than other SOTA methods. Specifically, compared with the norm or manifold metric-based methods, e.g., AFN and DMP, the OT metric ensures better characterization of data structure and achieves significantly higher accuracies. Moreover, compared with reweighing-based OT alignment, the OT alignment with importance sampling indeed achieves better knowledge transfer, i.e., IS$^2$C outperforms AR and m-POT.

Compared with Office-Home and VisDA-2017, transfer tasks on Office-31 are relatively simple and the performance is generally high. From the results in Table~\ref{tab:Office31}, we see IS$^2$C still improves the adaptation performance significantly even though the accuracies of SOTA methods are close to 100\%. Specifically, for the first 4 tasks, IS$^2$C achieves 100\% accuracies on the target domain, and the improvement on task A$\to$D exceeds 1.3\%. Similarly, compared with OT-based method (e.g., AR) and selection module-based method (e.g., SLM and SAN++), IS$^2$C learns latent information from the sampling data, which encourages the model to learn smaller risk on the source domain and further reduces the generalization risk on the target domain. These results validate our theoretical analysis.

\begin{table}[!t]
    \renewcommand{\arraystretch}{0.8}
    \begin{center}
        \caption{Classification accuracies ($ \% $) on Image-CLEF dataset (ResNet-50). }
        \label{tab:CLEF}
        \renewcommand{\tabcolsep}{0.38pc} 
        \begin{tabular}{c*{7}{c}}
            \toprule
            \textbf{Image-CLEF} &I$ \to $P	&P$ \to $I	&I$ \to $C	&C$ \to $I	&C$ \to $P	&P$ \to $C	& \textbf{Mean}	\\
            \midrule
            Baseline \cite{he2016deep}&78.3&86.9&91.0&84.3&72.5&91.5&84.1\\
            DANN \cite{ganin2016domain}&78.1&86.3&91.3&84.0&72.1&90.3&83.7\\
            PADA \cite{cao2018partial}&81.7&92.1&94.6&89.8&77.7&94.1&88.3\\
            HAFN \cite{xu2019larger}&79.1&87.7&93.7&90.3&77.8&94.7&87.2\\
            SAFN \cite{xu2019larger}&79.5&90.7&93.0&90.3&77.8&94.0&87.5\\
            DMP \cite{luo2020unsupervised}&82.4 & \textbf{94.5} &96.7 &94.3 &78.7 &96.4 &90.5 \\
            MUL \cite{luo2022generalized}& \textbf{87.5} & 92.2 & \textbf{98.1} &94.6 &87.6 & \textbf{98.5} &93.1\\
            MOT~\cite{luo2023mot}&87.7 &95.0 &98.0 &95.0 &87.0 &98.7 &93.6\\
            \midrule
            IS$^2$C&87.3	& {92.7}	&98.0	& \textbf{95.7}	& \textbf{87.7}	&98.0	& \textbf{93.2}\\
            \bottomrule
        \end{tabular}
    \end{center}
\end{table}

The results on Image-CLEF dataset are shown in Table~\ref{tab:CLEF}. Since Image-CLEF is a class-balanced dataset, the label shift on shared class is less significant. In such a scenario, the conditional distribution alignment, which is crucial to preserve the discriminability on the target domain, will be more important. Therefore, the results in Table~\ref{tab:CLEF} show that the methods with conditional distribution matching, e.g., MUL and IS$^2$C, consistently outperform other SOTA methods. Specifically, compared with other SOTA methods that cannot mitigate conditional shift, IS$^2$C is 2.7\% higher in accuracy at least. Therefore, the overall results demonstrate that IS$^2$C can effectively reduce the distribution shifts on label space and feature space, and further improves the transfer performance on both simple and challenging datasets.

\begin{figure*}[t] \centering
    \subfloat[Target Accuracy]{
        \includegraphics[width=0.238\textwidth]{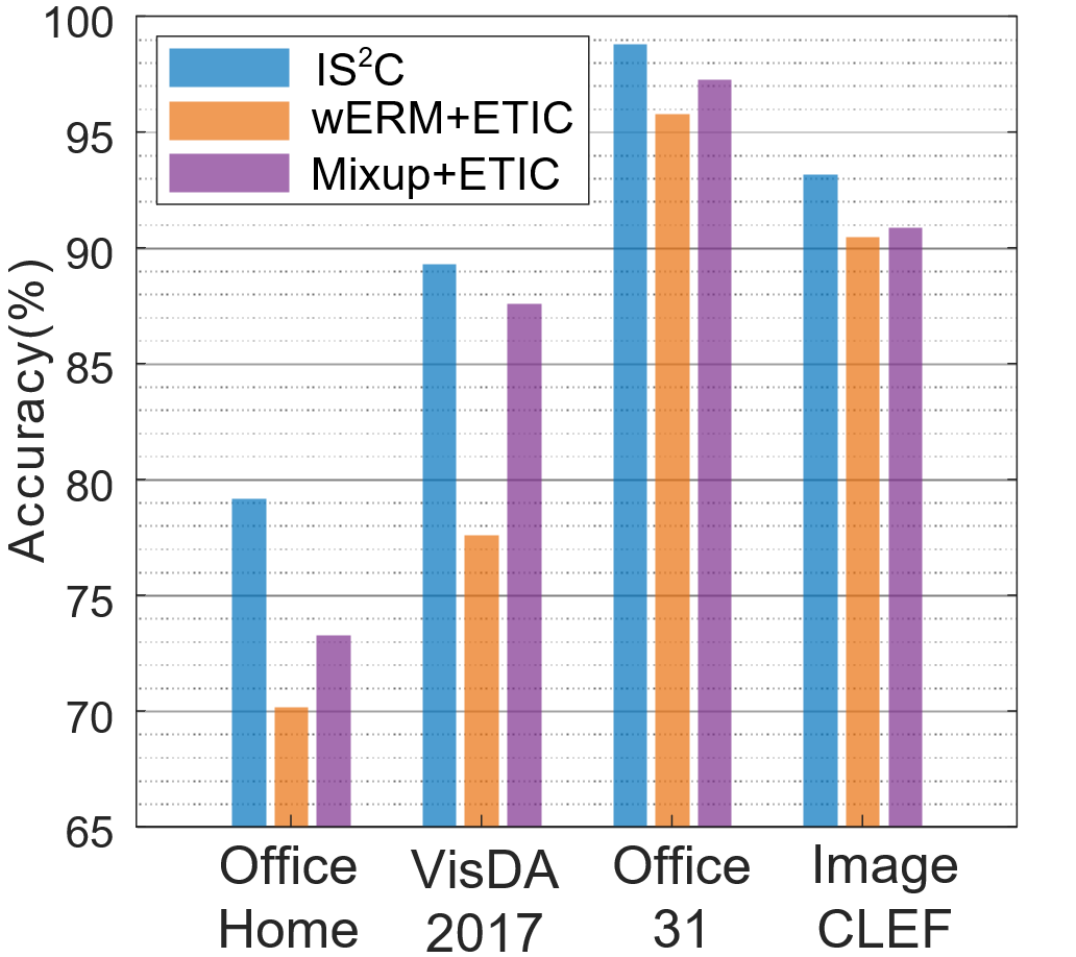}
        \label{fig:mixup_weighting_comparison}}
    \subfloat[$ \mathcal{A} $-distance]{
       \includegraphics[width=0.238\textwidth]{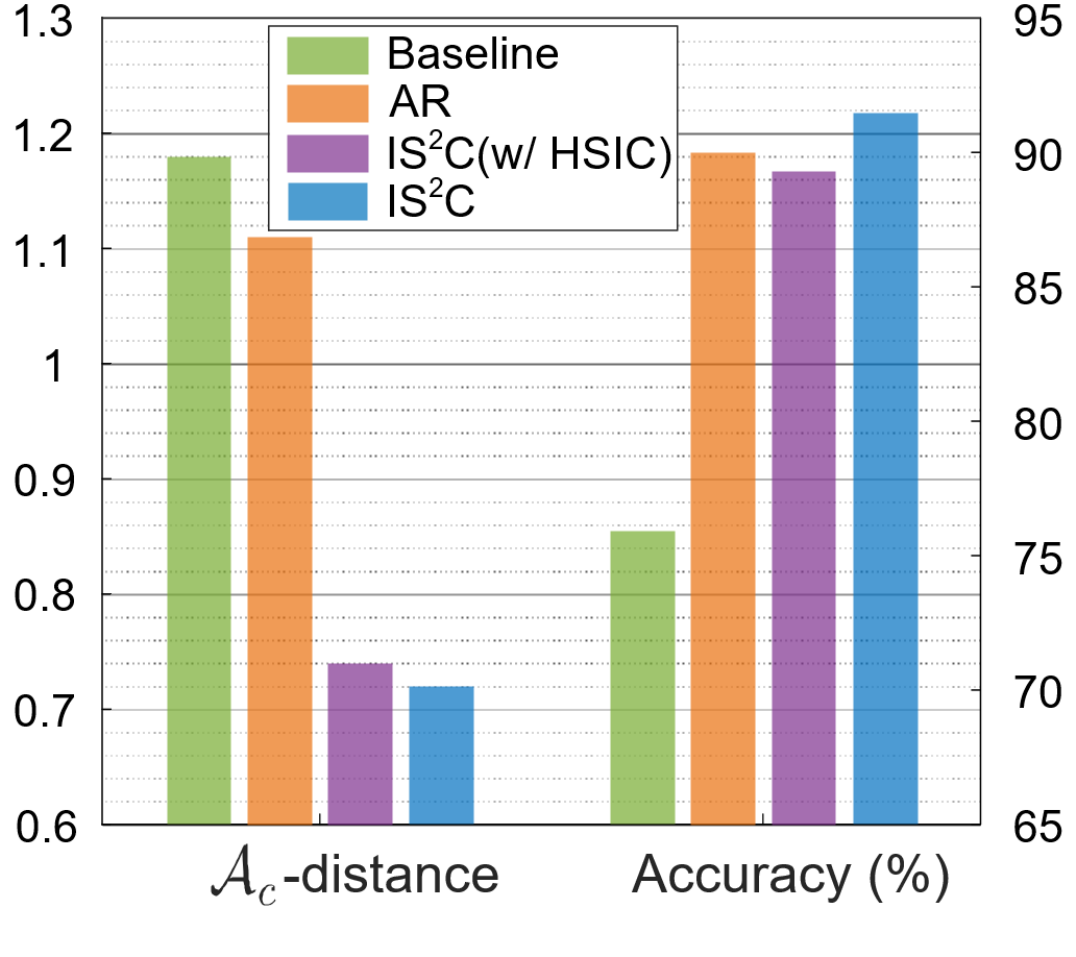}
        \label{fig:A_distance_bar}}
    \subfloat[Time Comparison]{
        \includegraphics[width=0.238\textwidth]{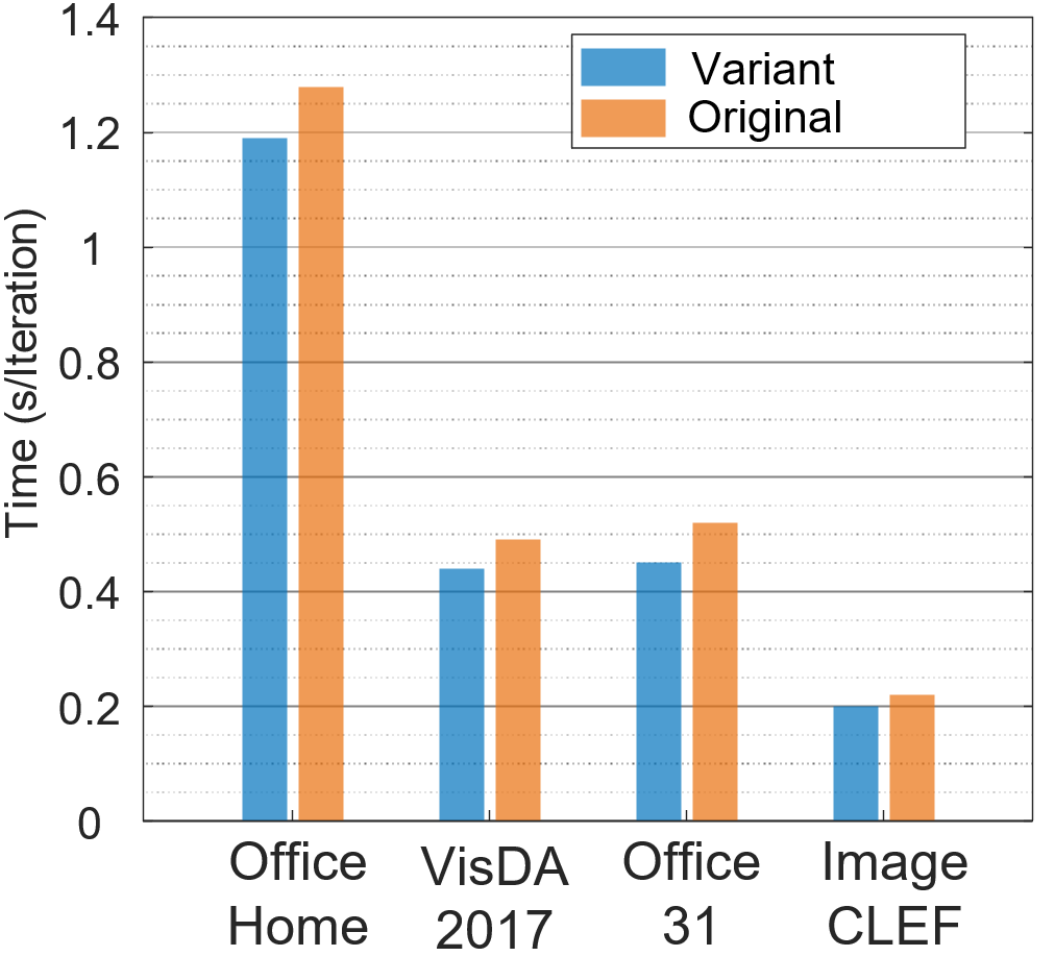}
        \label{fig:time_comparison}}
    \subfloat[Hyper-parameters]{       \includegraphics[width=0.238\textwidth]{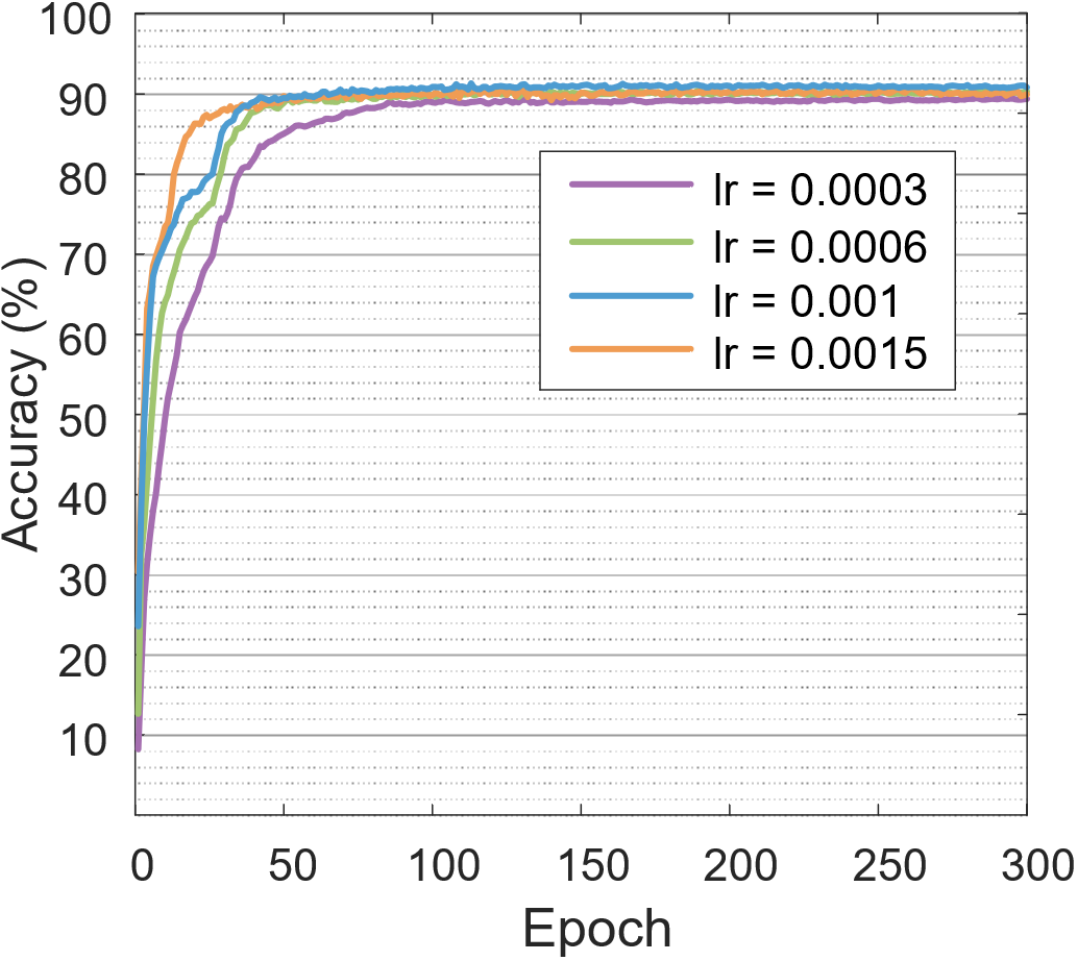}
\label{fig:sensitivity_home_ArRw_lr_epoch}}
   
    \caption{(a). Comparison between importance sampling strategies, importance weighting strategies, and mixup augmentation, which are denoted as IS$^2$C, wERM+ETIC and Mixup+ETIC, respectively. (b). Comparison of the domain discrepancies of different models, i.e., Baseline, AR, IS$^2$C (w/ HSIC) and IS$^2$C. (c). Comparison of the time cost between the original ETIC computation method and our proposed fast computation method. (d). Hyper-parameter analysis w.r.t. learning rate and the number of epochs.}
    \label{fig:IWIS_Adist_Time}
\end{figure*}

\begin{figure*}[!t] \centering
    \subfloat[$\mu$ and $\alpha$ (Ar$\to$Rw)]{
        \includegraphics[width=0.235\textwidth]{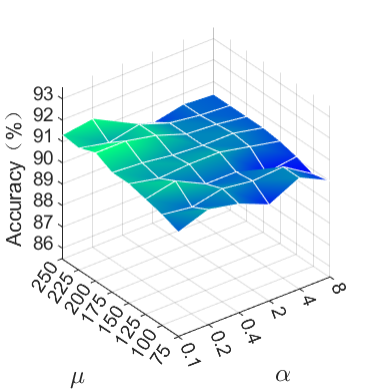}
        \label{fig:sensitivity_home_ArRw}}
    \subfloat[$\mu$ and $\alpha$ (W$\to$A)]{
     \includegraphics[width=0.235\textwidth]{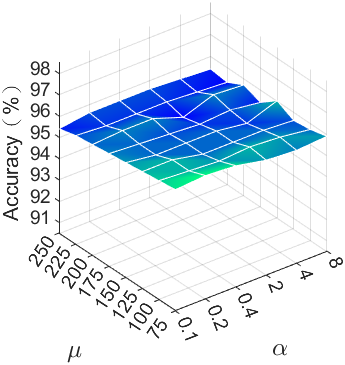}
        \label{fig:sensitivity_31_AW}}
    \subfloat[$\mu$ and $\theta$ (Ar$\to$Rw)]{
        \includegraphics[width=0.235\textwidth]{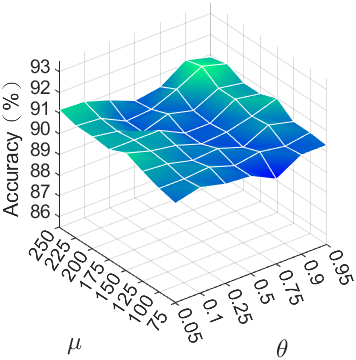}
        \label{fig:sensitivity_home_ArRw_directly}}
    \subfloat[$\mu$ and $\theta$ (W$\to$A)]{
     \includegraphics[width=0.235\textwidth]{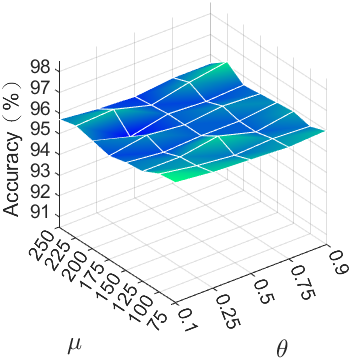}
        \label{fig:sensitivity_31_AW_directly}}
    \caption{Sensitivity analysis on hyper-parameters on Office-Home task Ar$\to$Rw and Office-31 task W$\to$A. (a)-(b). The results with different choices of balance parameter $\mu$ and sampling parameter $\alpha$. (c)-(d). The results with different choices of balance parameter $\mu$ and mix-ratio $\theta$.}
    \label{fig:Sensitivity}
\end{figure*}

\subsection{Other Analysis}

\begin{table}[!t]
    \renewcommand{\arraystretch}{0.8}
    \begin{center}
        \caption{Ablation study on different modules.}
        \label{tab:ablation}
        \begin{tabular}{*{7}{c}}
            \toprule
            \multicolumn{2}{c}{\textbf{Modules}} &Office&VisDA&Office&Image&\multirow{2}{*}{\textbf{Mean}}\\
            sampling&ETIC&Home&2017 &31&CLEF &	\\
            \midrule
            \XSolid&\XSolid&61.4&45.3&87.1&84.1&69.5\\
            \Checkmark&\XSolid&67.5&73.1&95.6&90.0&81.6\\
            \XSolid&\Checkmark&70.1&77.5&95.6&90.1&83.3\\
            \Checkmark &\Checkmark & \textbf{79.2}& \textbf{89.3}&\textbf{98.8}&\textbf{93.2}&\textbf{90.1}\\
            \bottomrule
        \end{tabular}
    \end{center}
\end{table}

\begin{figure*}[!t] \centering
    \subfloat[Synthetic Data($\sigma=10$)]{
        \includegraphics[width=0.235\textwidth]{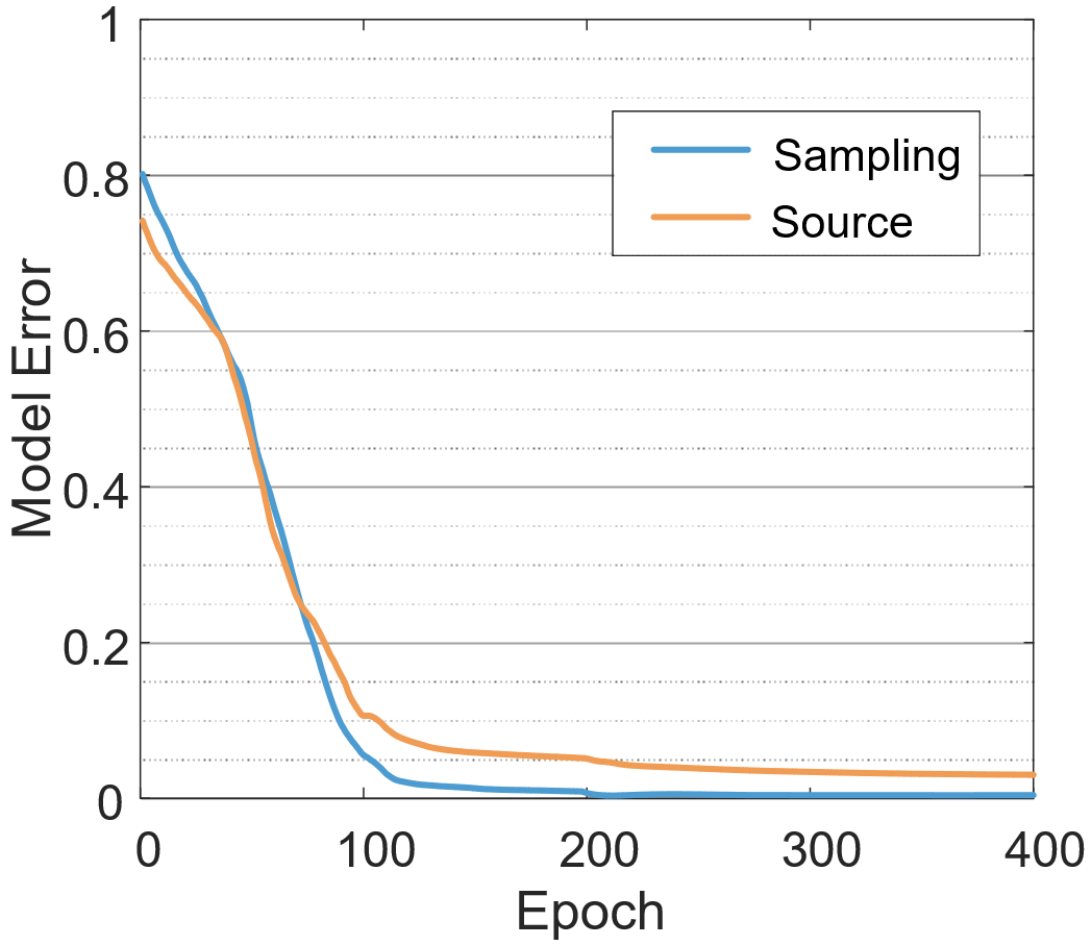}
        \label{fig:theta0point5_Ori_Mix_Error_Comparison}}
    \subfloat[Synthetic Data($\sigma=50$)]{
     \includegraphics[width=0.235\textwidth]{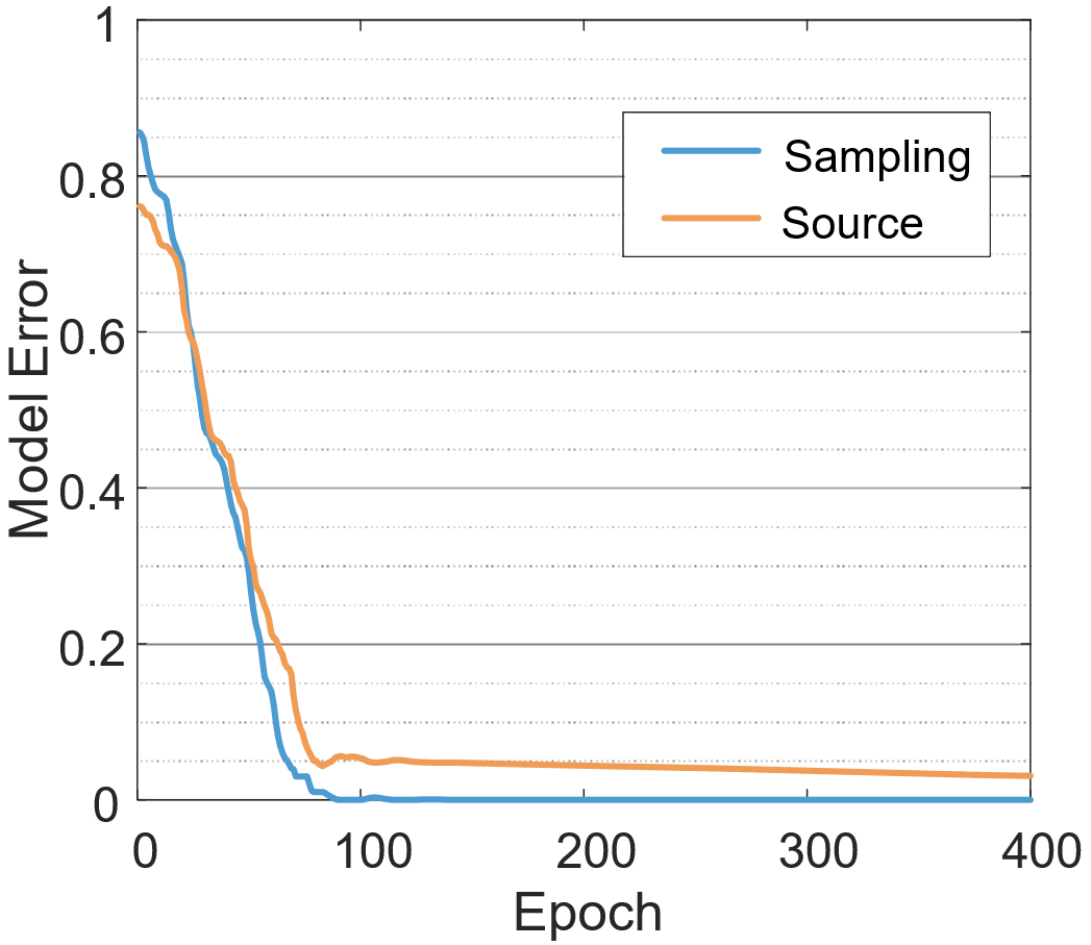}
        \label{fig:theta0point5_cor_50_Ori_Mix_Error_Comparison}}
    \subfloat[Art]{
        \includegraphics[width=0.235\textwidth]{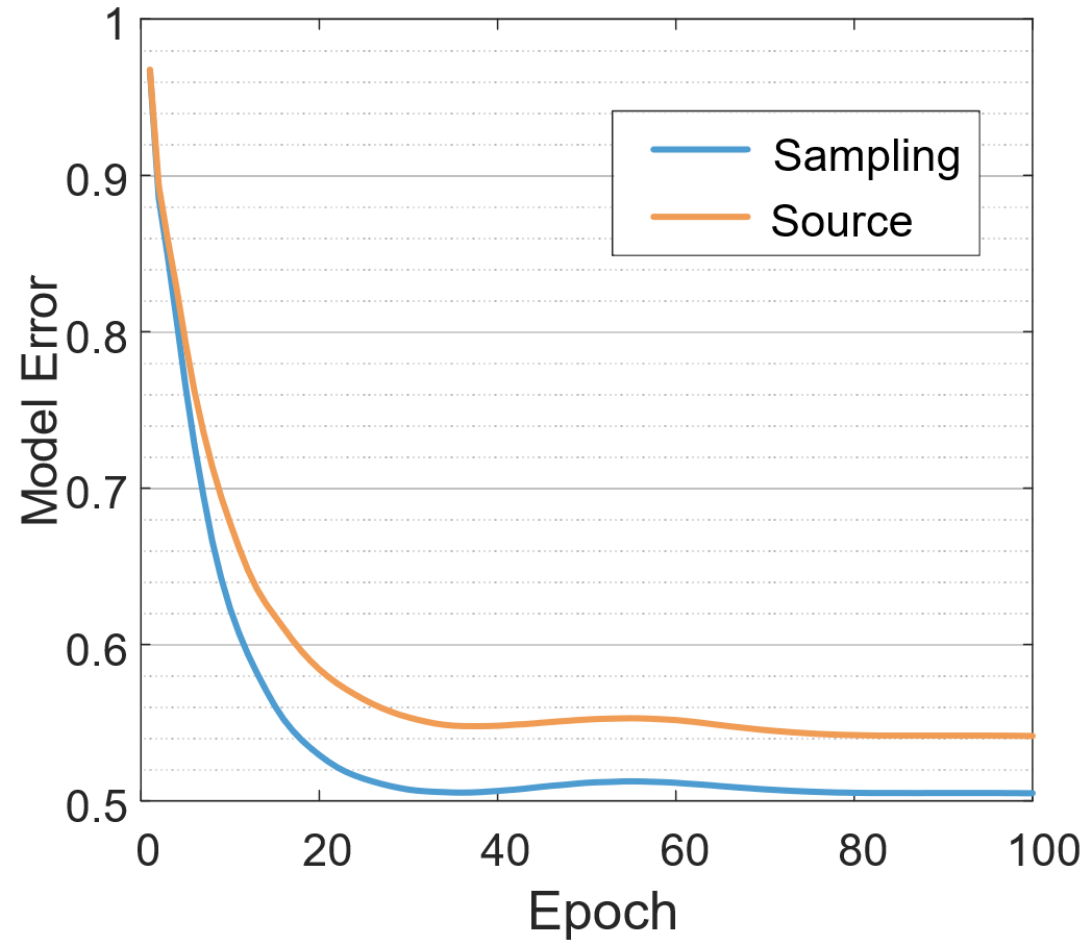}
        \label{fig:Art_Ori_Mix_Error_Comparison}}
    \subfloat[Webcam]{
        \includegraphics[width=0.235\textwidth]{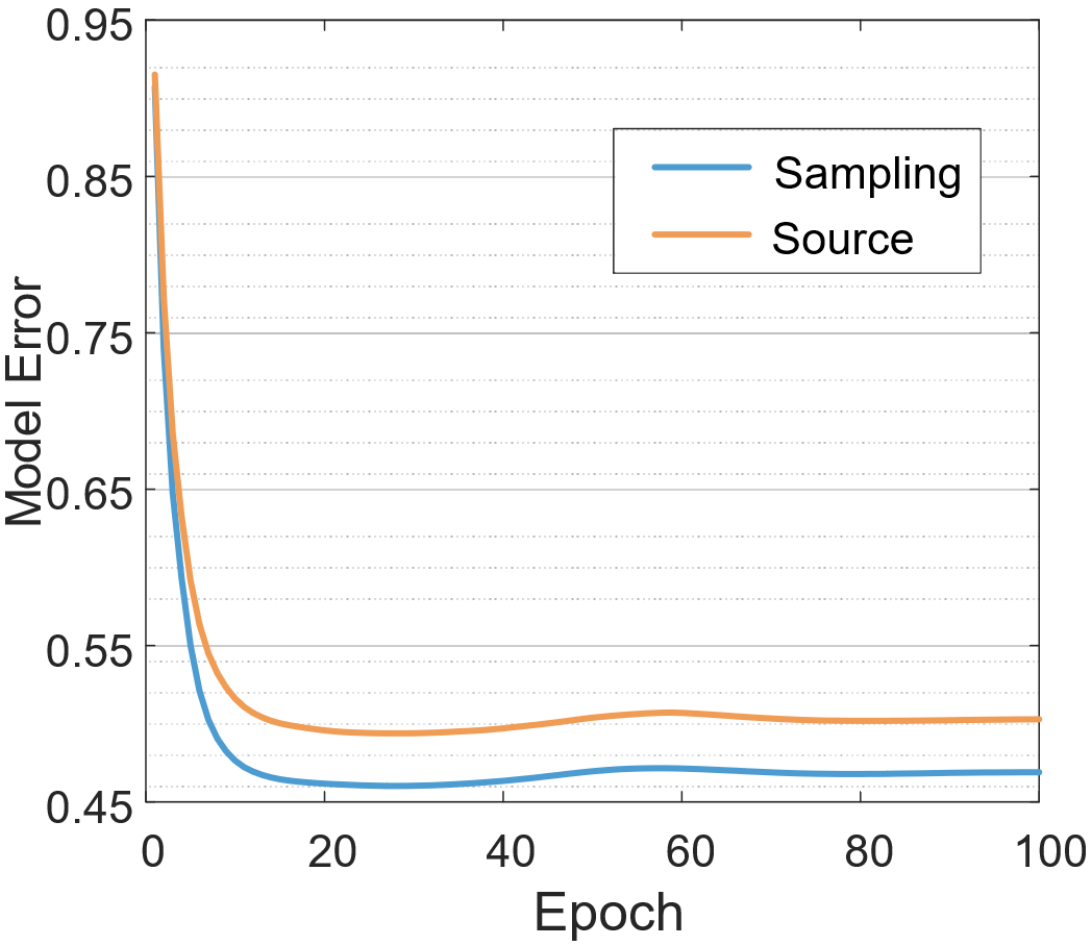}        \label{fig:Webcam_Ori_Mix_Error_Comparison}}
    \caption{Model error of $h \circ g$ over training epochs on the source and sampling domains. (a)-(b). Using synthetic data. (c)-(d). Using real data.}
    \label{fig:Ori_Mix_Error_Comparison}
\end{figure*}

\begin{figure*}[t] \centering
    \subfloat[Source]{
        \includegraphics[width=0.235\textwidth]{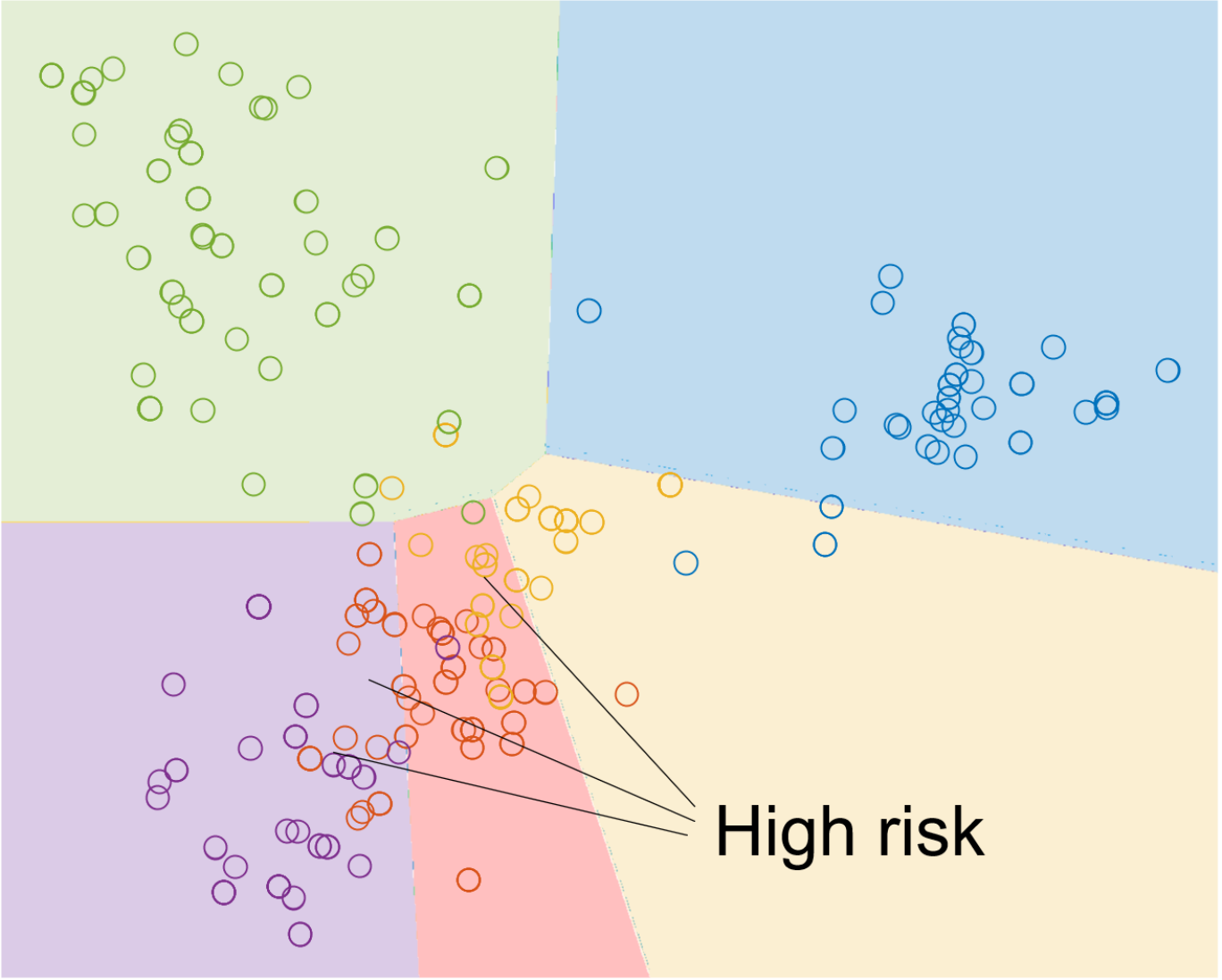}
        \label{fig:decision_boundary_source}}
    \subfloat[Sampling ($\theta=0.1$)]{
        \includegraphics[width=0.235\textwidth]{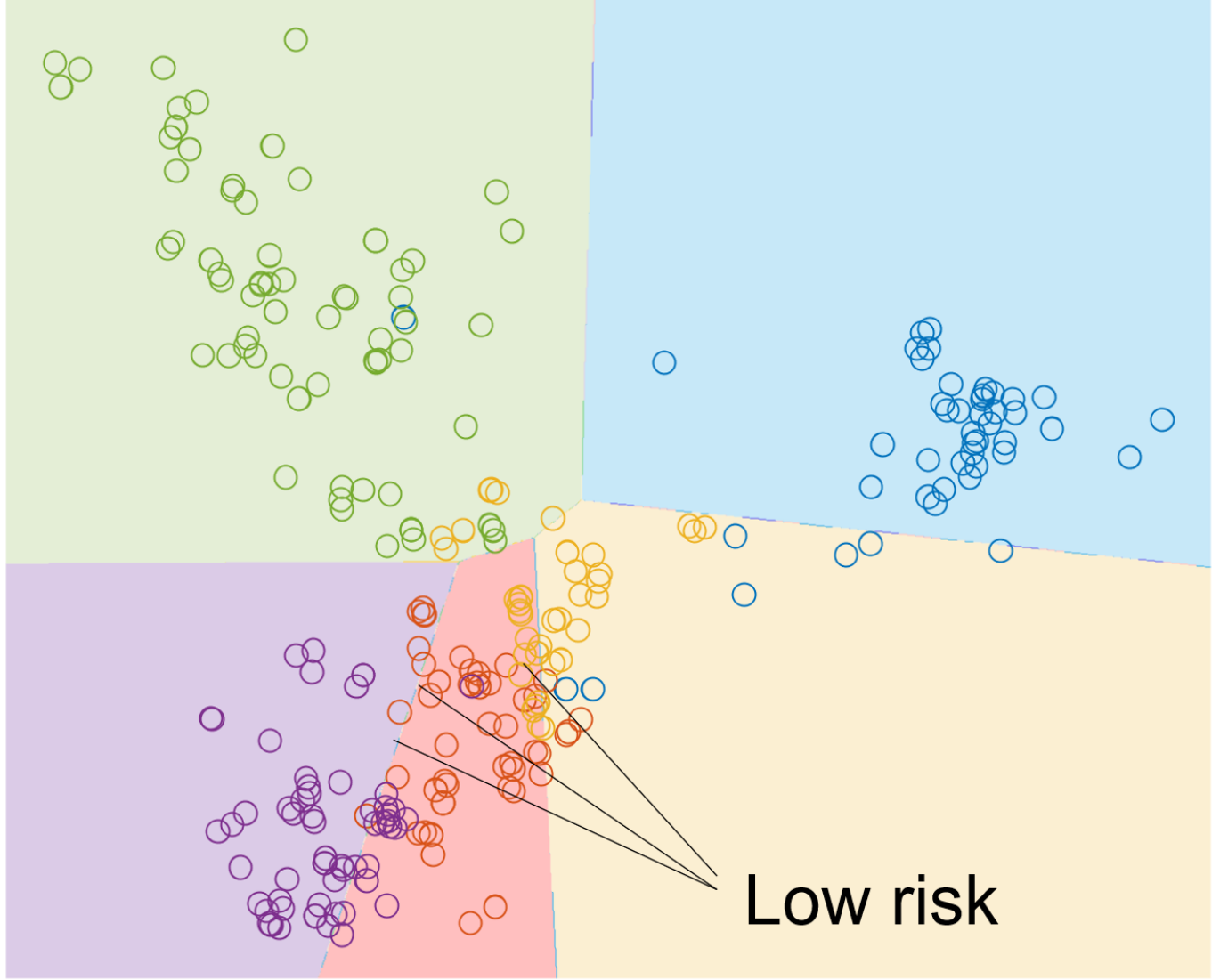}        \label{fig:decision_boundary_sampling_0.1}}
    \subfloat[Sampling ($\theta=0.25$)]{
        \includegraphics[width=0.235\textwidth]{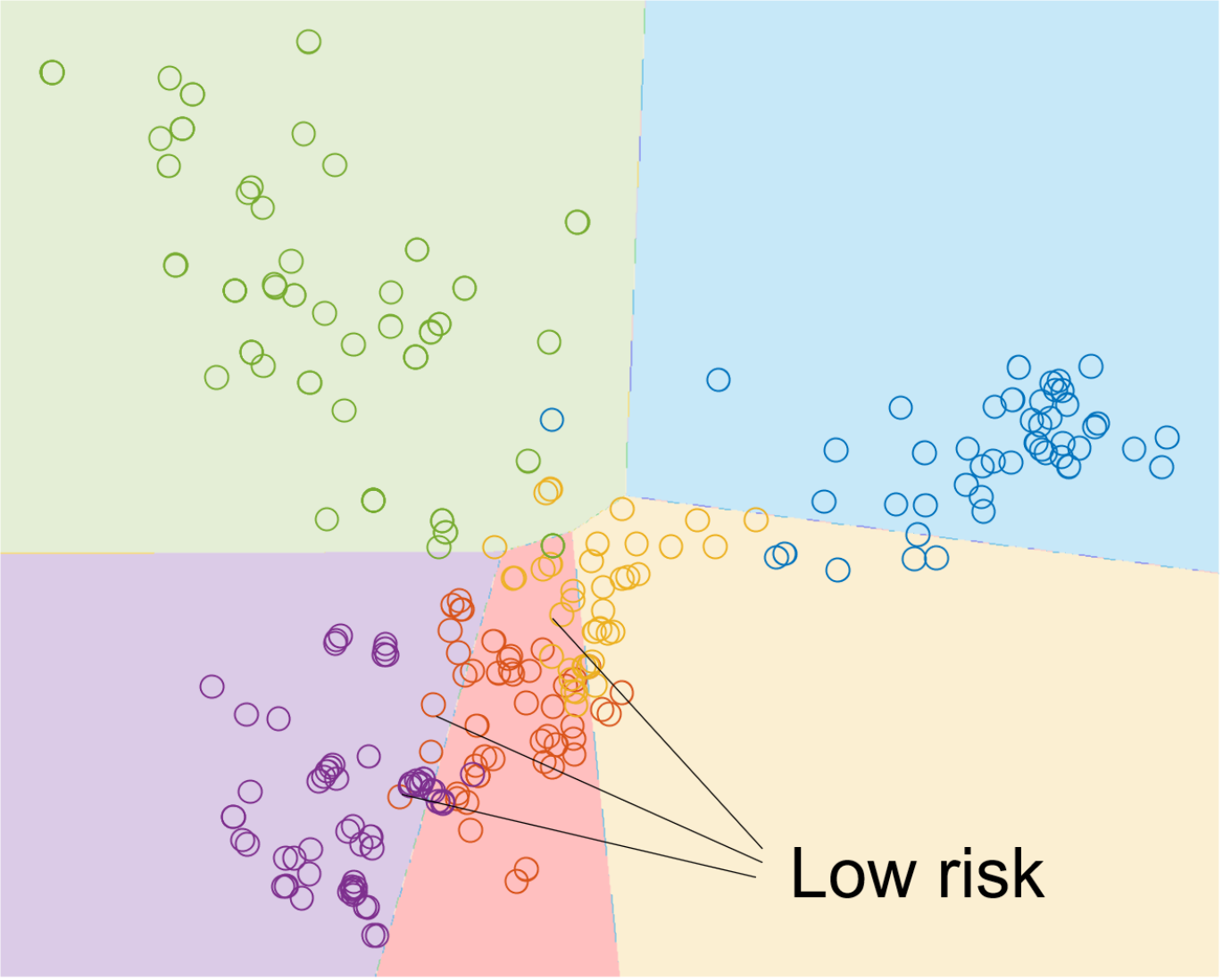}      \label{fig:decision_boundary_sampling_0.25}}   
    \subfloat[Sampling ($\theta=0.5$)]{
        \includegraphics[width=0.235\textwidth]{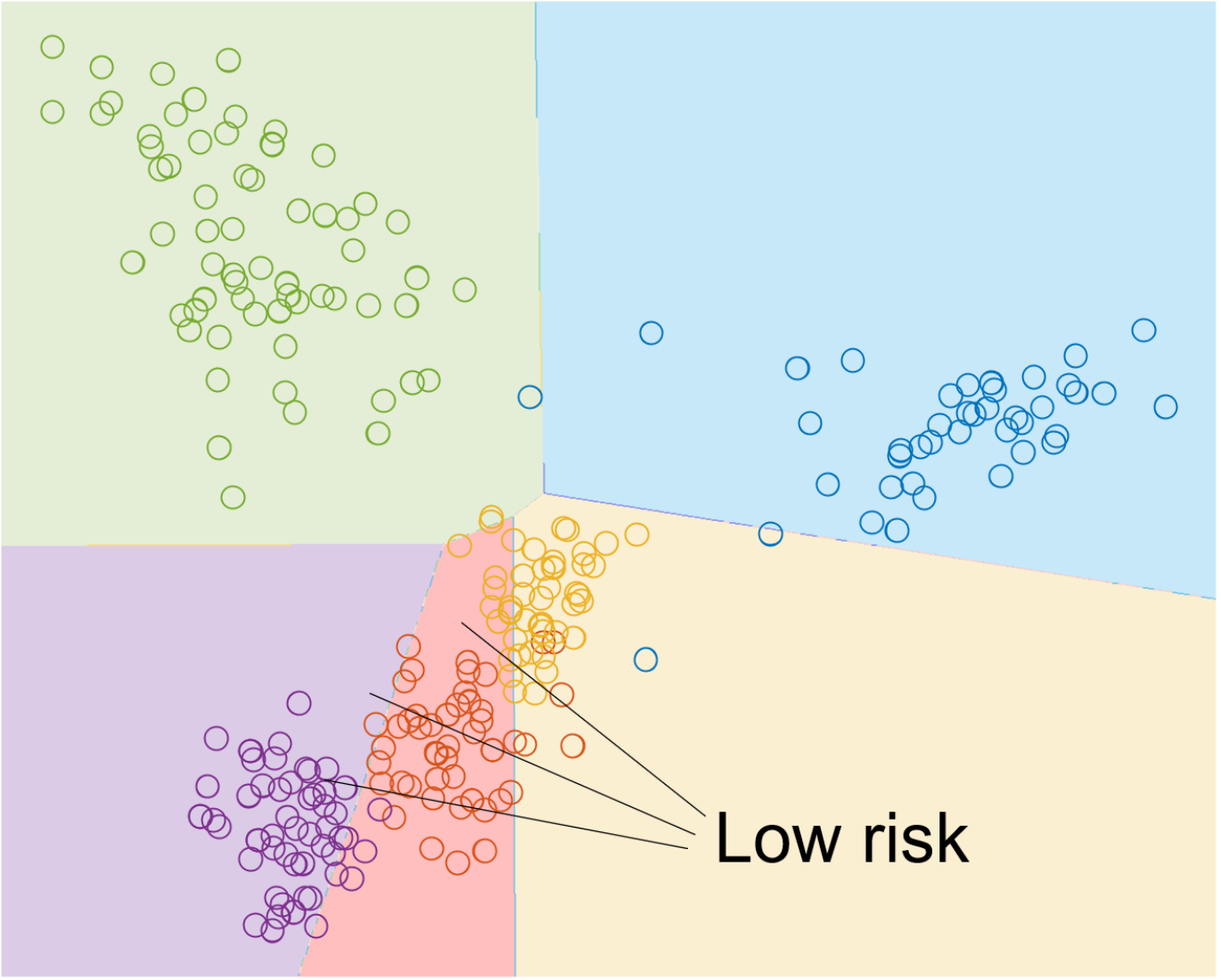}
        \label{fig:decision_boundary_sampling}}
    \caption{Two-dimensional visualizations for the source space, sampling space, and their decision boundaries on Office-Home dataset, where different colors represent different classes.}
    \label{fig:decision_boundary}
\end{figure*}

\begin{figure*}[t] \centering
    \subfloat[Source]{
        \includegraphics[width=0.32\textwidth]{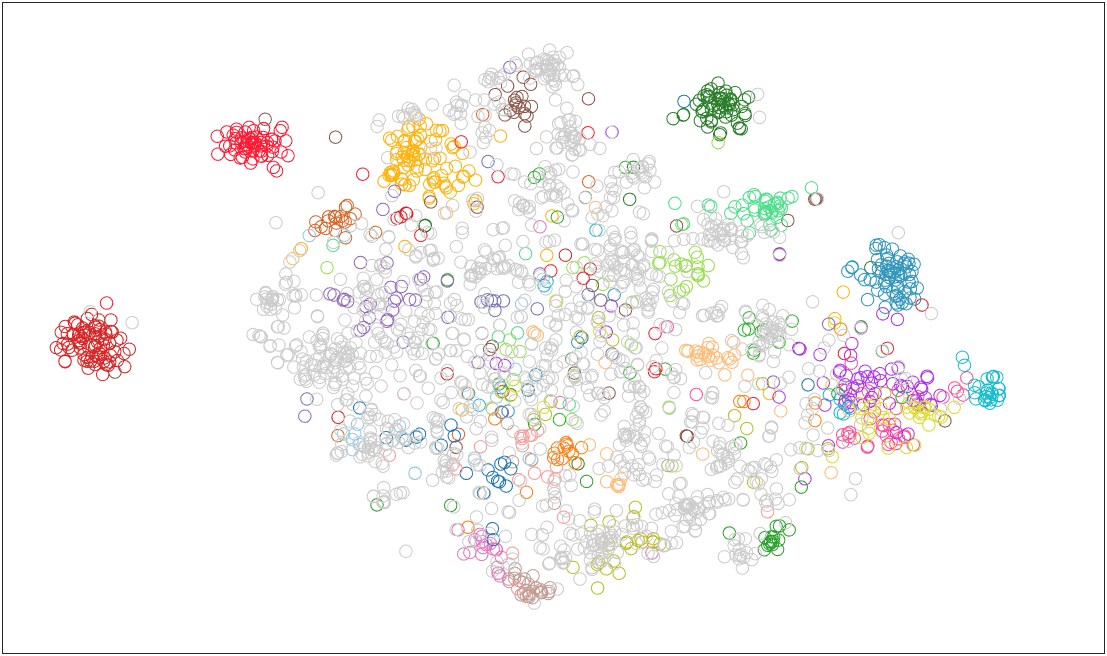}
        \label{fig:tSNE_source}}
    \subfloat[Sampling]{
        \includegraphics[width=0.32\textwidth]{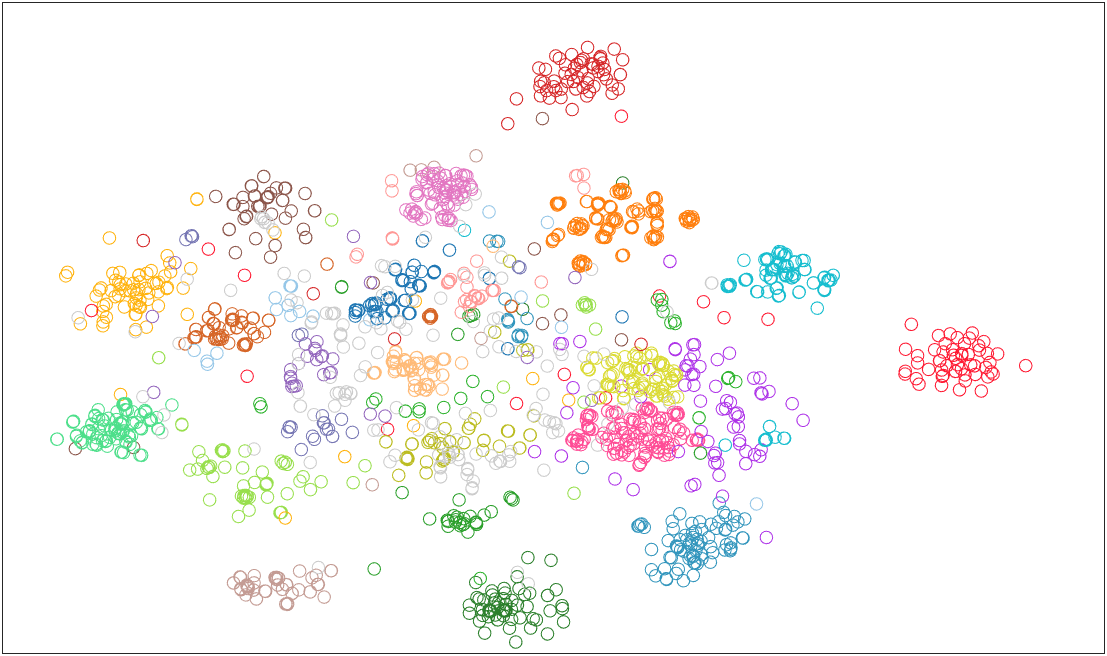}
        \label{fig:tSNE_sampling_source}}
    \subfloat[Baseline]{
        \includegraphics[width=0.32\textwidth]{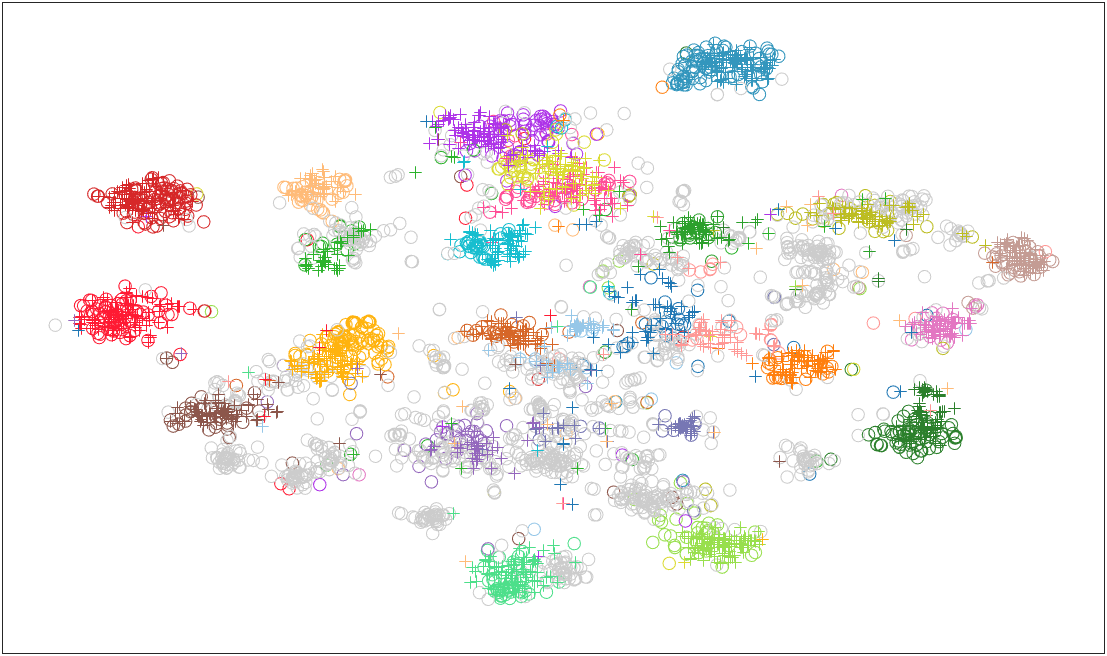}
        \label{fig:tSNE_baseline}}
    \\
    \subfloat[AR]{
        \includegraphics[width=0.32\textwidth]{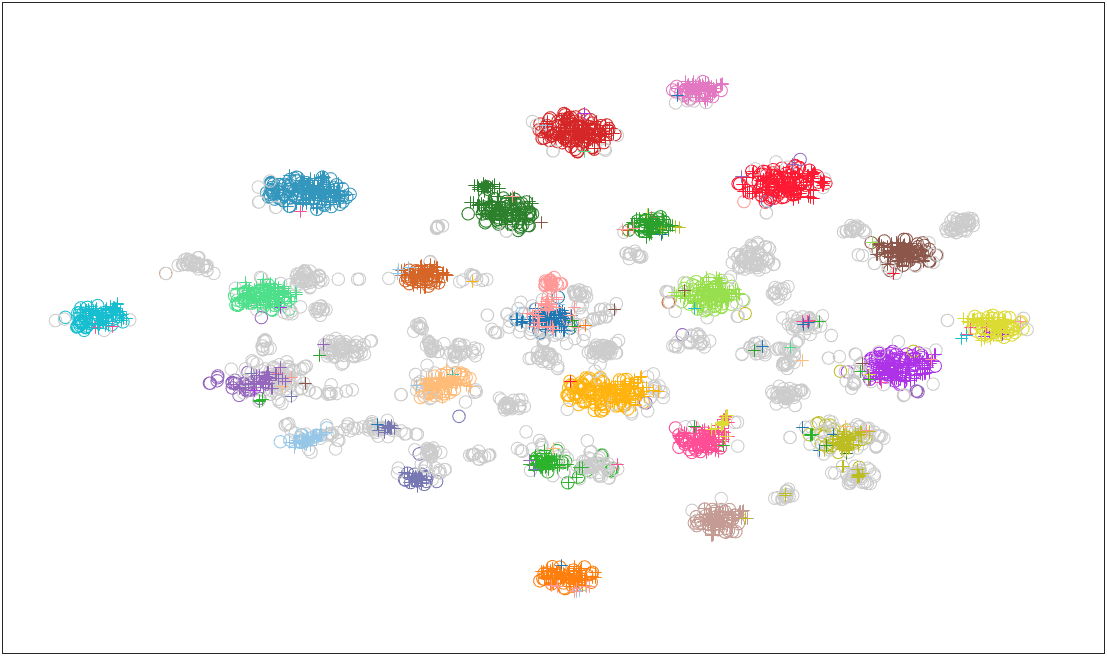}
        \label{fig:tSNE_AR}}
    \subfloat[IS$^2$C (w/ HSIC)]{
        \includegraphics[width=0.32\textwidth]{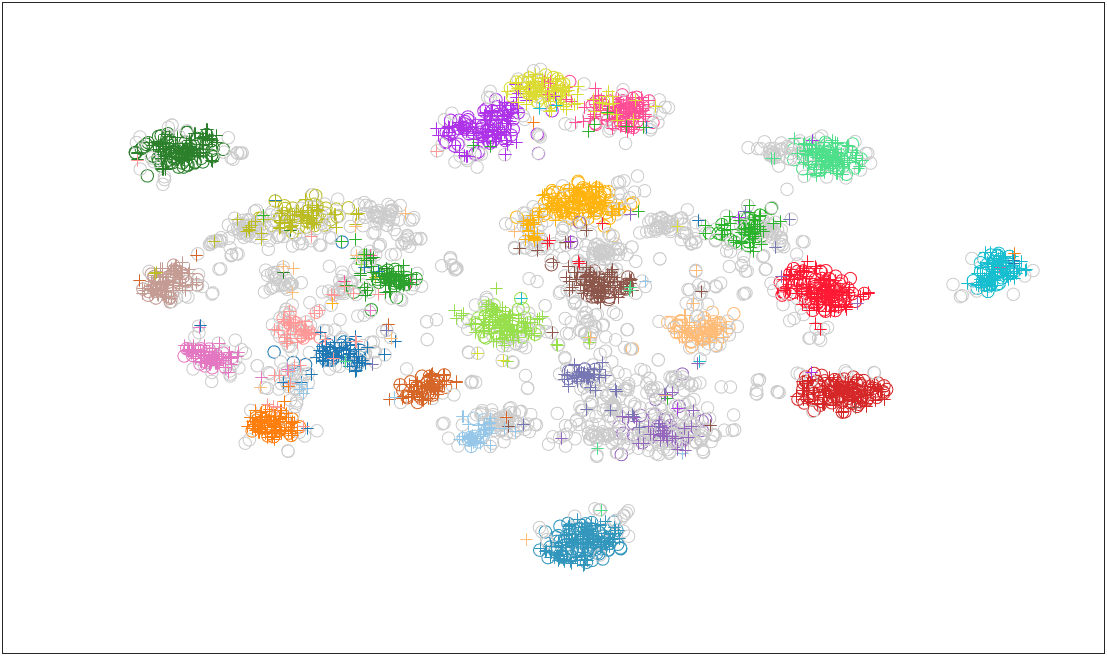}
        \label{fig:tSNE_MIS_HSIC}}
    \subfloat[IS$^2$C]{
        \includegraphics[width=0.32\textwidth]{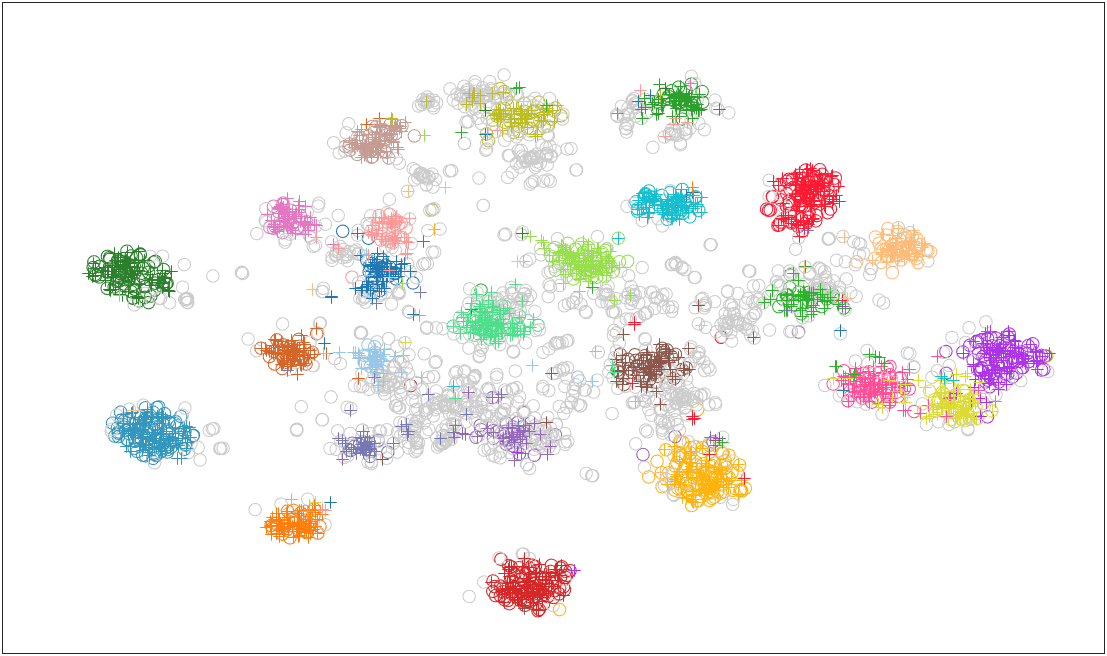}
        \label{fig:tSNE_MIS}}
    \caption{t-SNE \cite{van2008visualizing} visualizations of the representations learned by different PDA methods on the Office-Home task Ar$\to$Rw. In the diagrams, ``$\circ$'' means source or sampling domain, and ``+'' means target domain; different colors represent different classes, and gray represents outlier classes. (a)-(b). Samples from source domain and sampling domain. (c)-(f). The representations of the source and target samples learned by different PDA methods.}
    \label{fig:tSNE_visualization}
\end{figure*}

\textbf{Ablation Study.}
To evaluate the effectiveness of the proposed sampling module and ETIC-based alignment module, we conduct ablation experiments on 4 PDA datasets. The results are presented in Table~\ref{tab:ablation}. The first row and fourth row represent Baseline and IS$^2$C respectively. The second row is the result of Baseline model with sampling module, i.e., IS$^2$C without ETIC-based alignment; the third row is the result of ETIC only, i.e., without sampling module. As seen from Table~\ref{tab:ablation}, both the sampling module and ETIC module improve the accuracies of Baseline model on 4 datasets consistently, (about 12\%$\sim$14\% on average). Besides, in more challenging datasets, e.g., Office-Home and VisDA-2017, ETIC module achieves higher accuracies than the sampling module, which implies that the invariant representation learning via class-conditional alignment plays a crucial role in complex real-world scenarios. Furthermore, IS$^2$C consistently achieves the best performance on average $ 90.1\% $, which is significantly higher than the model with single module. Therefore, the sampling module and ETIC module can be benefited from each other, and then improve the adaptation performance significantly. In conclusion, the experiment results of the ablation study demonstrate that the proposed modules are generally effective in different datasets.

\textbf{The Effectiveness of IS.}
As discussed from methodology aspects, the IS strategy simultaneously mitigates the impact of
outlier classes and provides additional information from the unseen generated data, which ensures lower risk and better generalization. To validate the effectiveness of IS, we conduct experiments on 4 datasets to compare the target accuracies of IS-based model, IW-based model, and mixup-based model, denoted by IS$^2$C, wERM+ETIC, and Mixup+ETIC, respectively. Specifically, wERM+ETIC model replaces $\mathcal{L}_{risk}$ with the risk on reweighed source domain; following~\cite{na2022contrastive}, Mixup+ETIC model implements mixup~\cite{zhang2018mixup} augmentation with both source and target data (use pseudo label) to generate new mixup data without label shift correction, and replaces $\mathcal{L}_{risk}$ with the risk on the mixup data. From the results in Figure~\ref{fig:mixup_weighting_comparison}, we can observe that Mixup+ETIC outperforms wERM+ETIC on 4 datasets by an average of $3.8\%$, which shows that the creation of new unseen data plays a crucial role in improving model performance. Moreover, IS$^2$C model, i.e., label shift correction with IS, consistently outperforms wERM+ETIC model on 4 datasets. On Office-31 and Image-CLEF, IS-based model improves the performance by $ 3.0\% $ and $ 2.7\% $, respectively; on more challenging datasets, i.e., Office-Home and VisDA-2017, IS strategy significantly outperforms IW strategy by $ 9.0\% $ and $ 11.7\% $ in target accuracy, respectively. Besides, by introducing label shift correction, IS$^2$C consistently outperforms Mixup+ETIC by $5.9\%$, $1.7\%$, $1.5\%$, and $2.3\%$ on Office-Home, VisDA-2017, Office-31, and Image-CLEF, respectively. This indicates that the IS strategy has a significant advantage over conventional mixup augmentation in PDA tasks. In conclusion, the higher accuracies of IS$^2$C demonstrate the importance of label shift correction, and further show that the correction with sampling strategy is indeed superior to the existing weighting strategy.

\textbf{Class-conditional Discrepancy.}
To validate the effectiveness of IS$^2$C for representation learning, we conduct experiments to analyze the domain discrepancies of different models, i.e., Baseline, AR, IS$^2$C (w/ HSIC) and IS$^2$C; specifically, IS$^2$C (w/ HSIC) represent the ETIC is replaced by HSIC \cite{gretton2005kernel,gretton2007kernel}, which is a commonly used independence measure based on kernel methods, for distribution alignment. For discrepancy measure, the $ \mathcal{A} $-distance \cite{ben2010theory} is applied to the class-wise data to estimate the conditional discrepancy. The larger the distance, the larger the discrepancy. The experiment results on Office-Home task Ar$ \to $Rw are presented in Figure~\ref{fig:A_distance_bar}, where the class-conditional $ \mathcal{A}_c$-distance is computed as the mean of $ \mathcal{A} $-distances on all class-wise data. From the results, we can observe that the PDA methods significantly reduce the domain discrepancy of Baseline model; further, the class-conditional alignment methods, i.e., IS$^2$C (w/ HSIC) and IS$^2$C, generally ensure lower discrepancy and higher accuracy than the method with reweighing only, i.e., AR. Besides, the OT-based independence measure ETIC achieves better performance than the kernel-based measure HSIC. Overall, the results above validate that our method with IS and ETIC strategies consistently outperforms the existing strategies for shift correction.

\textbf{Time Comparison.}
To validate the effectiveness of the proposed ETIC adjusted computation method in reducing the time cost for PDA, we conduct experiments to analyze the time cost of each training iteration using different ETIC computation methods, i.e., Original and Variant; specifically, Original represents IS$^2$C training procedure with the original Tensor Sinkhorn algorithm to compute ETIC, while Variant represents replacing the original algorithm by the proposed computation method. For the time complexity measure, we record the time consumed each training iteration. The experiment results on four datasets are presented in Figure~\ref{fig:time_comparison}. Although the computation of ETIC is only a small part of the training procedure in each training iteration, we can still observe from the results that the variant method is consistently able to accelerate the calculation compared to the original method. Therefore, the lower time cost of the Variant demonstrates that the proposed adjusted computation method is indeed superior to the original one in practical PDA applications.

\textbf{Hyper-parameter Analysis.}
Since ETIC is an OT-based independence measure, to ensure a more accurate OT estimation, a larger batch-size is generally better. We train the model without mini-batch technique on relatively small datasets, including ImageCLEF, Office-31, and Office-Home. On VisDA-2017 dataset, we use the maximum batch size (5000) allowed by the GPU for mini-batch gradient descent. Besides, we also conduct experiments to know clear how sensitive the IS$^2$C model is to the learning rate (lr) and number of epochs. The results on Office-Home task Ar$\to$Rw are presented in Figure~\ref{fig:sensitivity_home_ArRw_lr_epoch}, apart from achieving the best performance at lr = 0.001, IS$^2$C also obtains high performance at other values of learning rate. The performance variation across different learning rates and numbers of epochs indicates the stability of IS$^2$C w.r.t. these two parameters.

Meanwhile, the IS$^2$C model is mainly related to two hyper-parameters, i.e., the balance parameter $\mu$ and the sampling parameter $\alpha$. The value of $\alpha$ essentially impacts the choice of $\theta$, thus we conduct experiments to investigate the sensitivity of the model w.r.t. $\mu$, $\alpha$ and $\theta$. The values of $\mu$, $\alpha$, and $\theta$ are selected from \{75, 100, 125, 150, 175, 200, 225, 250\}, \{0.1, 0.2, 0.4, 2, 4, 8\}, and \{0.05, 0.1, 0.25, 0.5, 0.75, 0.9, 0.95\}, respectively. The 3-D grid visualizations for the hyper-parameter sensitivity are presented in Figures~\ref{fig:sensitivity_home_ArRw}-\ref{fig:sensitivity_31_AW_directly}, where Office-Home task Ar$\to$Rw and Office-31 task W$\to$A are considered. According to the results, the IS$^2$C model is generally stable for different choices of hyper-parameters $\mu$. Moreover, as shown in Figures~\ref{fig:sensitivity_home_ArRw}-\ref{fig:sensitivity_31_AW}, the best performance is typically achieved when $\alpha$ is 0.2 or 0.4. Since $\alpha$ is a parameter for the beta distribution, $\alpha = 0.2$ or $0.4$ implies that in the sampling process, $\theta$ is more likely to take values close to 0 or 1. Thus, the results w.r.t. $\alpha$ indirectly suggest that $\theta$ close to 0 or 1 may contribute to better performance. Figures~\ref{fig:sensitivity_home_ArRw_directly}-\ref{fig:sensitivity_31_AW_directly} specifically show how different values of $\theta$ affect the model performance. We can observe that IS$^2$C is generally stable for different choices of $\theta$. Moreover, the best model performance is achieved at $\theta=0.1$ or $0.9$, while $\theta$ that are too close to $0.5$ slightly degrade the model performance on target domain. This validates our conclusion in Section \ref{sec3}, which states that when $\theta$ approximates $0.5$, it creates a sampling domain with overly compact cluster structure that negatively affects the model's generalization ability on the target domain. In conclusion, the results above demonstrate that IS$^2$C model is robust to the different settings of hyper-parameters and different learning environments, i.e., datasets.

\textbf{Risk of Sampling Domain.} As discussed in Section~\ref{sec3}, the classification risk on the sampling domain will be smaller than that on source domain. To empirically validate this conclusion, we first conduct experiments to visualize the representation spaces of source and sampling, where the decision boundaries are also provided. The experiment results on Office-Home task Ar$ \to $Rw are shown in Figure~\ref{fig:decision_boundary}, where the first 5 classes are presented. From the results, we observe that the IS strategy adjusts the decision boundary in sampling space according to different choices of mix-ratio $\theta$. Specifically, for the classes that are less discriminative and have a higher risk, i.e., purple, red, and yellow, the sampling space formed by IS strategy obtains a smaller risk value by encouraging the model to learn adjusted decision boundaries. Moreover, the sampling domain generated at $\theta=0.5$ does exhibit quite compact cluster structure mentioned in Section~\ref{sec3}. That means, although $\theta=0.5$ achieves a low risk on sampling domain, it may degrade the model's generalization ability on the target domain.   

Note that our method is motivated by the theoretical guidance of Proposition~\ref{pr1}. To support Proposition~\ref{pr1} under its convexity assumption, we conduct experiments on synthetic data using a convex model. Specifically, we generate source data by drawing 200 two-dimensional samples from $\mathcal{N}(\bm{0}, \sigma^2 \bm{I})$, and assign binary labels based on a linear decision boundary. Using the labeled source data for a given value of $\sigma$, we construct a corresponding sampling domain and set up a binary classification task. The model $h \circ g$ is implemented as a linear transformation followed by a ReLU activation and a clamp function, ensuring convexity and outputting a valid probability scalar in $[0,1]$. We train the model using cross-entropy on half of the source data, then test the model error on source/sampling domain using the full source/sampling data. The results for different values of $\sigma$ (where $\sigma\in \{10, 50\}$) are shown in Figures~\ref{fig:theta0point5_Ori_Mix_Error_Comparison}-\ref{fig:theta0point5_cor_50_Ori_Mix_Error_Comparison}. As training progresses, the model consistently exhibits lower error on the sampling domain than on the source domain. Moreover, the gap becomes more pronounced when source distribution has larger variance, aligning well with our expectations. These results provide empirical support for Proposition~\ref{pr1} when model convexity is satisfied.

However, the convexity assumption is quite strong and may limit its application in real-world scenarios, thus it is essential to reconsider Proposition~\ref{pr1} under the non-convex situation. Intuitively, we believe that the model error of non-convex $h \circ g$ on the sampling domain should still be smaller than its error on the source domain. We conduct experiments to empirically validate this conclusion. Specifically, we use the cross-entropy on half of the source data to train our $h\circ g$ (not convex, implemented as described in Section~\ref{sec4}) with varying numbers of epochs, then we test the model error of $h\circ g$ on source/sampling domain using the full source/sampling data. Figures~\ref{fig:Art_Ori_Mix_Error_Comparison}-\ref{fig:Webcam_Ori_Mix_Error_Comparison} presents the results on Art and Webcam, which are the source domains for task Ar$\to$Rw and task W$\to$A, respectively. According to Figures~\ref{fig:Art_Ori_Mix_Error_Comparison}-\ref{fig:Webcam_Ori_Mix_Error_Comparison}, although $h\circ g$ does not satisfy the convexity assumption, as the number of epochs increases, the model error on sampling domain remains smaller than that on source domain. This implies that while the convexity assumption of model $h\circ g$ in Proposition~\ref{pr1} is a sufficient condition, it may not be necessary, i.e., it can be relaxed in practical applications.   

In conclusion, these results ensure that the sampling with mixture distribution can help reduce the generalization error on both theoretical and empirical aspects.

\textbf{Feature visualization.} To further illustrate the role of importance sampling and show that our IS$^2$C does achieve class-conditional alignment for shifting distributions on shared classes, we visualize the learned representations of different PDA methods via t-SNE~\cite{van2008visualizing}.

The two-dimensional visualization results on Office-Home task Ar$\to$Rw are shown in Figure~\ref{fig:tSNE_visualization}. Figures \ref{fig:tSNE_source}-\ref{fig:tSNE_sampling_source} represent samples from source domain and sampling domain. The results imply that the IS strategy reduces samples of outlier classes in the sampling space, and ensures a better cluster structure, which are helpful in decreasing the model's risk. Further, the comparison for representations of different methods is shown in Figures~\ref{fig:tSNE_baseline}-\ref{fig:tSNE_MIS}. For the Baseline method in Figure~\ref{fig:tSNE_baseline}, the discriminability for shared classes is not sufficiently learned, and the negative impacts of outlier classes are still significant. By introducing label shift correction, IS$^2$C (w/ HSIC) and AR improve the identifiability between shared classes and outlier classes. However, note that there are still misalignment clusters in Figure~\ref{fig:tSNE_AR}, which demonstrates that the class conditional alignment in IS$^2$C indeed ensures better knowledge transfer. Besides, though IS$^2$C (w/ HSIC) also tries to align cross-domain clusters with HSIC measure, the more discriminative and compact structure of IS$^2$C implies that the OT-based ETIC measure does ensure better representation space with low risk and improved inter-class separation. In conclusion, these results validate the superiority of IS$^2$C in domain-adaptive representation learning.

\section{Conclusion}\label{sec6}

In this paper, we consider the limitations of mainstream methods for PDA, and propose a sampling-based method IS$^2$C to deal with the weaknesses. IS$^2$C is theoretically proved to admit appealing properties: 1) the generalization error is explicitly connected to the gap between source and sampling domains and can be sufficiently dominated under the IS$^2$C; 2) the new domain sampled via IS$^2$C has strictly smaller risk. Under the guarantees of theoretical results, a practical method for PDA is proposed by learning conditional independence and minimizing risk on sampling domain. Besides, an adjusted computation with quadratic complexity is explored for PDA applications, which overcomes the cubic complexity for computing independence criterion. Extensive experimental observations are consistent with theoretical results and demonstrate better performance compared with existing PDA methods.

How to extend the IS$^2$C to deal with open-set domain adaptation is our future work.

\bibliographystyle{IEEEtran}
\bibliography{mis-bib}

\begin{thebibliography}{10}
\providecommand{\url}[1]{#1}
\csname url@samestyle\endcsname
\providecommand{\newblock}{\relax}
\providecommand{\bibinfo}[2]{#2}
\providecommand{\BIBentrySTDinterwordspacing}{\spaceskip=0pt\relax}
\providecommand{\BIBentryALTinterwordstretchfactor}{4}
\providecommand{\BIBentryALTinterwordspacing}{\spaceskip=\fontdimen2\font plus
\BIBentryALTinterwordstretchfactor\fontdimen3\font minus
  \fontdimen4\font\relax}
\providecommand{\BIBforeignlanguage}[2]{{%
\expandafter\ifx\csname l@#1\endcsname\relax
\typeout{** WARNING: IEEEtran.bst: No hyphenation pattern has been}%
\typeout{** loaded for the language `#1'. Using the pattern for}%
\typeout{** the default language instead.}%
\else
\language=\csname l@#1\endcsname
\fi
#2}}
\providecommand{\BIBdecl}{\relax}
\BIBdecl

\bibitem{courty2017optimal}
N.~Courty, R.~Flamary, D.~Tuia \emph{et~al.}, ``Optimal transport for domain
  adaptation,'' \emph{IEEE Transactions on Pattern Analysis and Machine
  Intelligence}, vol.~39, no.~9, pp. 1853--1865, 2017.

\bibitem{pan2010domain}
S.~J. Pan, I.~W. Tsang, J.~T. Kwok \emph{et~al.}, ``Domain adaptation via
  transfer component analysis,'' \emph{IEEE Transactions on Neural Networks},
  vol.~22, no.~2, pp. 199--210, 2010.

\bibitem{liang2021SHOT}
J.~Liang, D.~Hu, Y.~Wang \emph{et~al.}, ``Source data-absent unsupervised
  domain adaptation through hypothesis transfer and labeling transfer,''
  \emph{IEEE Transactions on Pattern Analysis and Machine Intelligence},
  vol.~44, no.~11, pp. 8602--8617, 2021.

\bibitem{ben2010theory}
S.~Ben~David, J.~Blitzer, K.~Crammer \emph{et~al.}, ``A theory of learning from
  different domains,'' \emph{Machine Learning}, vol.~79, pp. 151--175, 2010.

\bibitem{david2010impossibility}
S.~Ben~David, T.~Lu, T.~Luu \emph{et~al.}, ``Impossibility theorems for domain
  adaptation,'' in \emph{Proceedings of the Thirteenth International Conference
  on Artificial Intelligence and Statistics}.\hskip 1em plus 0.5em minus
  0.4em\relax JMLR Workshop and Conference Proceedings, 2010, pp. 129--136.

\bibitem{borgwardt2006integrating}
K.~M. Borgwardt, A.~Gretton, M.~J. Rasch \emph{et~al.}, ``Integrating
  structured biological data by kernel maximum mean discrepancy,''
  \emph{Bioinformatics}, vol.~22, no.~14, pp. e49--e57, 2006.

\bibitem{long2015learning}
M.~Long, Y.~Cao, J.~Wang \emph{et~al.}, ``Learning transferable features with
  deep adaptation networks,'' in \emph{International Conference on Machine
  Learning}.\hskip 1em plus 0.5em minus 0.4em\relax PMLR, 2015, pp. 97--105.

\bibitem{zhu2020DSAN}
Y.~Zhu, F.~Zhuang, J.~Wang \emph{et~al.}, ``Deep subdomain adaptation network
  for image classification,'' \emph{IEEE Transactions on Neural Networks and
  Learning Systems}, vol.~32, no.~4, pp. 1713--1722, 2020.

\bibitem{sun2016return}
B.~Sun, J.~Feng, and K.~Saenko, ``Return of frustratingly easy domain
  adaptation,'' in \emph{Proceedings of the AAAI Conference on Artificial
  Intelligence}, vol.~30, no.~1, 2016, pp. 2058--2065.

\bibitem{gong2012geodesic}
B.~Gong, Y.~Shi, F.~Sha \emph{et~al.}, ``Geodesic flow kernel for unsupervised
  domain adaptation,'' in \emph{2012 IEEE Conference on Computer Vision and
  Pattern Recognition}.\hskip 1em plus 0.5em minus 0.4em\relax IEEE, 2012, pp.
  2066--2073.

\bibitem{ren2019heterogeneous}
C.~X. Ren, J.~Feng, D.~Q. Dai \emph{et~al.}, ``Heterogeneous domain adaptation
  via covariance structured feature translators,'' \emph{IEEE Transactions on
  Cybernetics}, vol.~51, no.~4, pp. 2166--2177, 2019.

\bibitem{luo2020unsupervised}
Y.~W. Luo, C.~X. Ren, D.~Q. Dai \emph{et~al.}, ``Unsupervised domain adaptation
  via discriminative manifold propagation,'' \emph{IEEE Transactions on Pattern
  Analysis and Machine Intelligence}, vol.~44, no.~3, pp. 1653--1669, 2020.

\bibitem{ganin2016domain}
Y.~Ganin, E.~Ustinova, H.~Ajakan \emph{et~al.}, ``Domain-adversarial training
  of neural networks,'' \emph{The Journal of Machine Learning Research},
  vol.~17, no.~1, pp. 2096--2030, 2016.

\bibitem{tzeng2017adversarial}
E.~Tzeng, J.~Hoffman, K.~Saenko \emph{et~al.}, ``Adversarial discriminative
  domain adaptation,'' in \emph{Proceedings of the IEEE Conference on Computer
  Vision and Pattern Recognition}, 2017, pp. 7167--7176.

\bibitem{tang2020discriminative}
H.~Tang and K.~Jia, ``Discriminative adversarial domain adaptation,'' in
  \emph{Proceedings of the AAAI Conference on Artificial Intelligence},
  vol.~34, no.~04, 2020, pp. 5940--5947.

\bibitem{ren2019domain}
C.~X. Ren, B.~Liang, P.~Ge \emph{et~al.}, ``Domain adaptive person
  re-identification via camera style generation and label propagation,''
  \emph{IEEE Transactions on Information Forensics and Security}, vol.~15, pp.
  1290--1302, 2019.

\bibitem{courty2017joint}
N.~Courty, R.~Flamary, A.~Habrard \emph{et~al.}, ``Joint distribution optimal
  transportation for domain adaptation,'' \emph{Advances in Neural Information
  Processing Systems}, vol.~30, pp. 3733--3742, 2017.

\bibitem{damodaran2018deepjdot}
B.~B. Damodaran, B.~Kellenberger, R.~Flamary \emph{et~al.}, ``Deepjdot: Deep
  joint distribution optimal transport for unsupervised domain adaptation,'' in
  \emph{Proceedings of the European Conference on Computer Vision (ECCV)},
  2018, pp. 447--463.

\bibitem{zhang2019optimal}
Z.~Zhang, M.~Wang, and A.~Nehorai, ``Optimal transport in reproducing kernel
  hilbert spaces: Theory and applications,'' \emph{IEEE Transactions on Pattern
  Analysis and Machine Intelligence}, vol.~42, no.~7, pp. 1741--1754, 2019.

\bibitem{ren2018generalized}
C.~X. Ren, X.~L. Xu, and H.~Yan, ``Generalized conditional domain adaptation: A
  causal perspective with low-rank translators,'' \emph{IEEE Transactions on
  Cybernetics}, vol.~50, no.~2, pp. 821--834, 2018.

\bibitem{kirchmeyer2021mapping}
M.~Kirchmeyer, A.~Rakotomamonjy, E.~de~Bezenac \emph{et~al.}, ``Mapping
  conditional distributions for domain adaptation under generalized target
  shift,'' \emph{arXiv preprint arXiv:2110.15057}, 2021.

\bibitem{tachet2020domain}
R.~Tachet~des Combes, H.~Zhao, Y.~X. Wang \emph{et~al.}, ``Domain adaptation
  with conditional distribution matching and generalized label shift,''
  \emph{Advances in Neural Information Processing Systems}, vol.~33, pp.
  19\,276--19\,289, 2020.

\bibitem{zhang2013domain}
K.~Zhang, B.~Sch{\"o}lkopf, K.~Muandet \emph{et~al.}, ``Domain adaptation under
  target and conditional shift,'' in \emph{International Conference on Machine
  Learning}.\hskip 1em plus 0.5em minus 0.4em\relax PMLR, 2013, pp. 819--827.

\bibitem{luo2022generalized}
Y.~W. Luo and C.~X. Ren, ``Generalized label shift correction via minimum
  uncertainty principle: Theory and algorithm,'' \emph{arXiv preprint
  arXiv:2202.13043}, 2022.

\bibitem{cao2019learning}
Z.~Cao, K.~You, M.~Long \emph{et~al.}, ``Learning to transfer examples for
  partial domain adaptation,'' in \emph{Proceedings of the IEEE/CVF Conference
  On Computer Vision and Pattern Recognition}, 2019, pp. 2985--2994.

\bibitem{li2020deep}
S.~Li, C.~H. Liu, Q.~Lin \emph{et~al.}, ``Deep residual correction network for
  partial domain adaptation,'' \emph{IEEE Transactions on Pattern Analysis and
  Machine Intelligence}, vol.~43, no.~7, pp. 2329--2344, 2020.

\bibitem{yan2017mind}
H.~Yan, Y.~Ding, P.~Li \emph{et~al.}, ``Mind the class weight bias: Weighted
  maximum mean discrepancy for unsupervised domain adaptation,'' in
  \emph{Proceedings of the IEEE Conference on Computer Vision and Pattern
  Recognition}, 2017, pp. 2272--2281.

\bibitem{cao2018partial}
Z.~Cao, L.~Ma, M.~Long \emph{et~al.}, ``Partial adversarial domain
  adaptation,'' in \emph{Proceedings of the European Conference on Computer
  Vision (ECCV)}, 2018, pp. 135--150.

\bibitem{kim2020associative}
Y.~Kim, S.~Hong, S.~Yang \emph{et~al.}, ``Associative partial domain
  adaptation,'' \emph{arXiv preprint arXiv:2008.03111}, 2020.

\bibitem{ge2023unsupervised}
P.~Ge, C.~X. Ren, X.~L. Xu \emph{et~al.}, ``Unsupervised domain adaptation via
  deep conditional adaptation network,'' \emph{Pattern Recognition}, vol. 134,
  p. 109088, 2023.

\bibitem{long2018conditional}
M.~Long, Z.~Cao, J.~Wang \emph{et~al.}, ``Conditional adversarial domain
  adaptation,'' \emph{Advances in Neural Information Processing Systems},
  vol.~31, pp. 1647--1657, 2018.

\bibitem{xu2020adversarial}
M.~Xu, J.~Zhang, B.~Ni \emph{et~al.}, ``Adversarial domain adaptation with
  domain mixup,'' in \emph{Proceedings of the AAAI conference on Artificial
  Intelligence}, vol.~34, no.~04, 2020, pp. 6502--6509.

\bibitem{li2020enhanced}
M.~Li, Y.~M. Zhai, Y.~W. Luo \emph{et~al.}, ``Enhanced transport distance for
  unsupervised domain adaptation,'' in \emph{Proceedings of the IEEE/CVF
  Conference on Computer Vision and Pattern Recognition}, 2020, pp.
  13\,936--13\,944.

\bibitem{luo2021conditional}
Y.~W. Luo and C.~X. Ren, ``Conditional bures metric for domain adaptation,'' in
  \emph{Proceedings of the IEEE/CVF Conference on Computer Vision and Pattern
  Recognition}, 2021, pp. 13\,989--13\,998.

\bibitem{ren2022buresnet}
C.~X. Ren, Y.~W. Luo, and D.~Q. Dai, ``Buresnet: Conditional bures metric for
  transferable representation learning,'' \emph{IEEE Transactions on Pattern
  Analysis and Machine Intelligence}, vol.~45, no.~4, pp. 4198--4213, 2022.

\bibitem{xu2019larger}
R.~Xu, G.~Li, J.~Yang \emph{et~al.}, ``Larger norm more transferable: An
  adaptive feature norm approach for unsupervised domain adaptation,'' in
  \emph{Proceedings of the IEEE/CVF International Conference on Computer
  Vision}, 2019, pp. 1426--1435.

\bibitem{na2022contrastive}
J.~Na, D.~Han, H.~J. Chang \emph{et~al.}, ``Contrastive vicinal space for
  unsupervised domain adaptation,'' in \emph{European Conference on Computer
  Vision}.\hskip 1em plus 0.5em minus 0.4em\relax Springer, 2022, pp. 92--110.

\bibitem{gu2021adversarial}
X.~Gu, X.~Yu, J.~Sun \emph{et~al.}, ``Adversarial reweighting for partial
  domain adaptation,'' \emph{Advances in Neural Information Processing
  Systems}, vol.~34, pp. 14\,860--14\,872, 2021.

\bibitem{nguyen2022improving}
K.~Nguyen, D.~Nguyen, T.~Pham \emph{et~al.}, ``Improving mini-batch optimal
  transport via partial transportation,'' in \emph{International Conference on
  Machine Learning}.\hskip 1em plus 0.5em minus 0.4em\relax PMLR, 2022, pp.
  16\,656--16\,690.

\bibitem{luo2023mot}
Y.~W. Luo and C.~X. Ren, ``{MOT}: Masked optimal transport for partial domain
  adaptation,'' in \emph{Proceedings of the IEEE/CVF Conference on Computer
  Vision and Pattern Recognition}, 2023, pp. 3531--3540.

\bibitem{sahoo2023select}
A.~Sahoo, R.~Panda, R.~Feris \emph{et~al.}, ``Select, label, and mix: Learning
  discriminative invariant feature representations for partial domain
  adaptation,'' in \emph{Proceedings of the IEEE/CVF Winter Conference on
  Applications of Computer Vision}, 2023, pp. 4210--4219.

\bibitem{wu2022ran}
K.~Wu, M.~Wu, Z.~Chen,  \emph{et~al.}, ``Reinforced adaptation network for
  partial domain adaptation,'' \emph{IEEE Transactions on Circuits and Systems
  for Video Technology}, 2022.

\bibitem{cao2022san}
Z.~Cao, K.~You, Z.~Zhang,  \emph{et~al.}, ``From big to small: Adaptive
  learning to partial-set domains,'' \emph{IEEE Transactions on Pattern
  Analysis and Machine Intelligence}, vol.~45, no.~2, pp. 1766--1780, 2022.

\bibitem{li2022idsp}
W.~Li and S.~Chen, ``Partial domain adaptation without domain alignment,''
  \emph{IEEE Transactions on Pattern Analysis and Machine Intelligence}, 2022.

\bibitem{ma2024small}
Y.~Ma, X.~Yao, R.~Chen \emph{et~al.}, ``Small is beautiful: Compressing deep
  neural networks for partial domain adaptation,'' \emph{IEEE Transactions on
  Neural Networks and Learning Systems}, vol.~35, no.~3, pp. 3575--3585, 2024.

\bibitem{lin2022CI}
K.~Y. Lin, J.~Zhou, Y.~Qiu,  \emph{et~al.}, ``Adversarial partial domain
  adaptation by cycle inconsistency,'' in \emph{European Conference on Computer
  Vision}.\hskip 1em plus 0.5em minus 0.4em\relax Springer, 2022, pp. 530--548.

\bibitem{tokdar2010importance}
S.~T. Tokdar and R.~E. Kass, ``Importance sampling: a review,'' \emph{Wiley
  Interdisciplinary Reviews: Computational Statistics}, vol.~2, no.~1, pp.
  54--60, 2010.

\bibitem{carratino2020mixup}
L.~Carratino, M.~Ciss{\'e}, R.~Jenatton \emph{et~al.}, ``On mixup
  regularization,'' \emph{arXiv preprint arXiv:2006.06049}, 2020.

\bibitem{liu2022entropy}
L.~Liu, S.~Pal, and Z.~Harchaoui, ``Entropy regularized optimal transport
  independence criterion,'' in \emph{International Conference on Artificial
  Intelligence and Statistics}.\hskip 1em plus 0.5em minus 0.4em\relax PMLR,
  2022, pp. 11\,247--11\,279.

\bibitem{gretton2005kernel}
A.~Gretton, R.~Herbrich, A.~Smola \emph{et~al.}, ``Kernel methods for measuring
  independence,'' \emph{Journal of Machine Learning Research}, vol.~6, no.~70,
  pp. 2075--2129, 2005.

\bibitem{gretton2007kernel}
A.~Gretton, K.~Fukumizu, C.~Teo \emph{et~al.}, ``A kernel statistical test of
  independence,'' \emph{Advances in Neural Information Processing Systems},
  vol.~20, 2007.

\bibitem{feydy2019interpolating}
J.~Feydy, T.~S{\'e}journ{\'e}, F.~X. Vialard \emph{et~al.}, ``Interpolating
  between optimal transport and mmd using sinkhorn divergences,'' in \emph{The
  22nd International Conference on Artificial Intelligence and
  Statistics}.\hskip 1em plus 0.5em minus 0.4em\relax PMLR, 2019, pp.
  2681--2690.

\bibitem{he2016deep}
K.~He, X.~Zhang, S.~Ren \emph{et~al.}, ``Deep residual learning for image
  recognition,'' in \emph{Proceedings of the IEEE Conference on Computer Vision
  and Pattern Recognition}, 2016, pp. 770--778.

\bibitem{lipton2018detecting}
Z.~Lipton, Y.~X. Wang, and A.~Smola, ``Detecting and correcting for label shift
  with black box predictors,'' in \emph{International Conference on Machine
  Learning}.\hskip 1em plus 0.5em minus 0.4em\relax PMLR, 2018, pp. 3122--3130.

\bibitem{deng2009imagenet}
J.~Deng, W.~Dong, R.~Socher,  \emph{et~al.}, ``Imagenet: A large-scale
  hierarchical image database,'' in \emph{IEEE Conference on Computer Vision
  and Pattern Recognition}.\hskip 1em plus 0.5em minus 0.4em\relax Ieee, 2009,
  pp. 248--255.

\bibitem{kingma2014adam}
D.~P. Kingma and J.~Ba, ``Adam: A method for stochastic optimization,''
  \emph{arXiv preprint arXiv:1412.6980}, 2014.

\bibitem{caputo2014imageclef}
B.~Caputo, H.~M{\"u}ller, J.~Martinez~Gomez \emph{et~al.}, ``Imageclef 2014:
  Overview and analysis of the results,'' in \emph{Information Access
  Evaluation. Multilinguality, Multimodality, and Interaction: 5th
  International Conference of the CLEF Initiative, CLEF 2014, Sheffield, UK,
  September 15-18, 2014. Proceedings 5}.\hskip 1em plus 0.5em minus 0.4em\relax
  Springer, 2014, pp. 192--211.

\bibitem{saenko2010adapting}
K.~Saenko, B.~Kulis, M.~Fritz \emph{et~al.}, ``Adapting visual category models
  to new domains,'' in \emph{Computer Vision--ECCV 2010: 11th European
  Conference on Computer Vision, Heraklion, Crete, Greece, September 5-11,
  2010, Proceedings, Part IV 11}.\hskip 1em plus 0.5em minus 0.4em\relax
  Springer, 2010, pp. 213--226.

\bibitem{venkateswara2017deep}
H.~Venkateswara, J.~Eusebio, S.~Chakraborty \emph{et~al.}, ``Deep hashing
  network for unsupervised domain adaptation,'' in \emph{Proceedings of the
  IEEE Conference on Computer Vision and Pattern Recognition}, 2017, pp.
  5018--5027.

\bibitem{peng2017visda}
X.~Peng, B.~Usman, N.~Kaushik \emph{et~al.}, ``Visda: The visual domain
  adaptation challenge,'' \emph{arXiv preprint arXiv:1710.06924}, 2017.

\bibitem{van2008visualizing}
L.~Van~der Maaten and G.~Hinton, ``Visualizing data using t-sne.''
  \emph{Journal of Machine Learning Research}, vol.~9, no.~11, 2008.

\bibitem{zhang2018mixup}
H.~Zhang, ``mixup: Beyond empirical risk minimization,'' in \emph{International
  Conference on Learning Representations}, 2018, pp. 1--13.

\end{thebibliography}

\end{document}